\documentclass{article}

\usepackage{graphicx,amscd,amsmath,amssymb,verbatim}
\usepackage[dvips]{hyperref}
\usepackage[TS1,OT1,T1]{fontenc}

\begin{document}

\title{Adaptive Cluster Expansion (ACE): A Hierarchical Bayesian
Network\footnote{This paper was submitted to IEEE Trans. PAMI on
2 May 1991. Paper PAMI No. 91-05-18. It was not accepted for publication,
but it underpins several subsequently published papers.}}
\author{Stephen Luttrell}
\maketitle

\noindent {\bfseries Abstract:} Using the maximum entropy method,
we derive the ``adaptive cluster expansion'' (ACE), which can be
trained to estimate probability density functions in high dimensional
spaces. The main advantage of ACE over other Bayesian networks is
its ability to capture high order statistics after short training
times, which it achieves by making use of a hierarchical vector
quantisation of the input data. We derive a scheme for representing
the state of an ACE network as a ``probability image'', which allows
us to identify statistically anomalous regions in an otherwise statistically
homogeneous image, for instance. Finally, we present some probability
images that we obtained after training ACE on some Brodatz texture
images - these demonstrate the ability of ACE to detect subtle textural
anomalies.
\section{Introduction}

The purpose of this paper is to train probabilistic network models
of images of homogeneous textures for use in Bayesian decision making.
In our past work in this area \cite{{Luttrell1987b, Luttrell1987c,
Luttrell1987d, Luttrell1987a, Luttrell1988c}} we successfully used
entropic methods to design Markov random field (MRF) models to reproduce
the observed statistical properties of textured images. We now wish
to formulate a novel MRF structure that requires much less effort
to train and use. There are two essential ingredients in our simplification:
we do not use hidden variables, and we restrict our attention to
hierarchical transformations of the data.

The use of hidden variables is a flexible way of modelling high
order correlations in data \cite{AckleyHintonSejnowski1985}, but
it leads to lengthy Monte Carlo simulations to estimate averages
over the hidden variables. An MRF without hidden variables is specified
by a set of transformation functions, each of which extracts some
statistic from the data, and together they provide sufficient information
to compute the probability density function (PDF) of the data \cite{{Luttrell1987a,
Luttrell1988c}}.

We can obtain a wealth of statistical information about the data
by restricting our attention to a finite number of well-defined
transformation functions. For instance, in \cite{HaralickShanmugamDinstein1973}
a number of useful textural features are presented, which may be
used to model and discriminate between various textures that occur
in images. However, we wish to design our transformation functions
adaptively in a data-driven manner, so that the resulting set is
optimised to capture the statistical properties of the data. We
choose to use adaptive hierarchical transformation functions, because
these not only capture statistical properties at many length scales,
but are also very easy to train.

We briefly discussed hierarchical transformation functions in \cite{Luttrell1989d},
where we conjectured that topographic mappings \cite{Kohonen1984}
might be appropriate for connecting together the layers of the hierarchy.
We investigated topographic mappings in \cite{{Luttrell1988a, Luttrell1988b,
Luttrell1989a, Luttrell1989e, Luttrell1990b}} and found that they
could be rapidly trained to produce useful multiscale representations
of data. We therefore use multilayer topographic mappings to adaptively
design hierarchical transformation functions of data for use in
MRF models. In this type of model different layers of the hierarchy
measure statistical structure on different length scales, and shorter
length scale structures are clustered together and correlated to
produce longer length scale structures. We therefore frequently
refer to this type of scheme as an adaptive cluster expansion (ACE).
By interpreting ACE as a multilayered {\itshape n}-tuple processor
we can relate ACE to a multilayered version of WISARD \cite{AleksanderStonham1979}.

We demonstrate the ability of ACE to learn the statistical structure
of texture by training an adaptive pyramid image processor. There
are many ways of displaying the statistical information extracted
from the data by such a processor, but we prefer to use what we
call a ``probability image'', which is generated from the estimated
local PDF of the data.

The layout of this paper is as follows. In Section \ref{XRef-Section-104222534}
we use the maximum entropy method to estimate the PDF of the data,
subject to a set of marginal probability constraints measured using
hierarchical transformation functions, to yield an MRF model in
closed form (i.e. no undetermined Lagrange multipliers). In Section
\ref{XRef-Section-104222545} we extend this result to remove some
of the limitations of its hierarchical structure, such a translation
non-invariance, and describe the ACE system for producing probability
images. In Section \ref{XRef-Section-104222552} we present the result
of applying ACE to some textured images taken from the Brodatz set
\cite{Brodatz1966}.
\section{Maximum entropy PDF estimation}\label{XRef-Section-104222534}

In this section we present a derivation of a hierachical maximum
entropy estimate $Q_{\textup{mem}}( x) $ of an observed true PDF
$P( x) $, where we constrain $Q_{\textup{mem}}( x) $ so that certain
marginal PDFs agree with observation. Although we consider only
the case of a binary tree, we also present a simple diagrammatic
representation of this result that allows us easily to extend it
to general trees.
\subsection{Basic maximum entropy method}

For completeness, we first of all outline the basic principles \cite{{Jaynes1968,
Jaynes1982}} of the maximum entropy method of assigning estimates
of PDFs. Introduce the entropy functional $H$
\begin{equation}
H=-\int dx Q( x) \log ( \frac{Q( x) }{Q_{0}( x) })
\end{equation}
in which the PDF $Q_{0}( x) $ is used to introduce prior knowledge
about $P( x) $. Loosely speaking, $H$ measures the extent to which
$Q( x) $ is non-committal about the value that $x$ might take. The
maximum entropy method consists of maximising $H$ subject to the
following set of constraints
\begin{equation}
\begin{array}{rl}
 C_{1,i} & =\int dx Q( x) y_{i}( x) -\int dx P( x) y_{i}( x)  \\
  & =0
\end{array}%
\label{XRef-Equation-104223834}
\end{equation}
where the $y_{i}( x) $ are the components of a vector $y( x) $ of
sampling functions. These constraints ensure that certain average
values are the same whether they are measured using $Q( x) $ (i.e.
our estimated PDF) or using $P( x) $ (i.e. the observed true PDF).
By carefully selecting the $y( x) $ we can optimise the agreement
between $Q( x) $ and $P( x) $ as appropriate.

$Q_{\textup{mem}}( x) $ may be found by introducing a vector $\lambda
$ of Lagrange multipliers, and functionally differentiating $H-\sum
_{i}\lambda _{i}C_{1,i}$ with respect to $Q( x) $ to yield eventually
\begin{equation}
Q_{\textup{mem}}( x) =\frac{Q_{0}( x) \exp ( -\lambda .y( x) ) }{\int
dx^{\prime }Q_{0}( x^{\prime }) \exp ( -\lambda .y( x^{\prime })
) }%
\label{XRef-Equation-104224355}
\end{equation}
The undetermined Lagrange vector $\lambda $ must be chosen in such
a way that the constraints are satisfied - this is usually a non-trivial
problem.

Now we shall consider a special case of the maximum entropy problem
in which we carefully design the $y_{i}( x) $ so that they constrain
a set of marginal probabilities \cite{Luttrell1988c}. Thus we make
the following replacements
\begin{equation}
\begin{array}{rl}
 y_{i}( x)  & \longrightarrow \delta ( y-y( x) )  \\
 \lambda _{i} & \longrightarrow \lambda ( y)
\end{array}%
\label{XRef-Equation-104223825}
\end{equation}
where $\delta ( y-y( x) ) $ is a Dirac delta function. In the $\{y_{i}(
x) ,\lambda _{i}\}$ version of the maximum entropy problem, by varying
the value of an index $i$ we could scan through the set of constraint
functions $y_{i}( x) $ and Lagrange multipliers $\lambda _{i}$.
However, in the $\{\delta ( y-y( x) ) ,\lambda ( y) \}$ version
of the maximum entropy problem, by varying the value of a variable
$y$ we can scan through the set of constraint functions $\delta
( y-y( x) ) $ and Lagrange multipliers $\lambda ( y) $.

The modification in Equation \ref{XRef-Equation-104223825} causes
the constraints in Equation \ref{XRef-Equation-104223834} to become
\begin{equation}
\begin{array}{rl}
 C_{2}( y)  & \equiv \int dx Q( x) \delta ( y-y( x) ) -\int dx P(
x) \delta ( y-y( x) )  \\
  & =Q( y) -P( y)  \\
  & =0
\end{array}%
\label{XRef-Equation-10422420}
\end{equation}
where we have defined the PDFs over $y$ as
\begin{equation}
\begin{array}{rl}
 Q( y)  & =\int dx Q( x) \delta ( y-y( x) )  \\
 P( y)  & =\int dx P( x) \delta ( y-y( x) )
\end{array}%
\label{XRef-Equation-104224212}
\end{equation}
Thus the delta function constraints force $Q( y) =P( y) $. Note
that we have used a rather loose notation for our PDFs - $P( x)
$ and $P( y) $ are in fact different functions of their respective
arguments. We have made this choice of notation for simplicity,
because the context will always indicate unambiguously which PDF
is required.

By analogy with the previous maximum entropy derivation, $Q_{\textup{mem}}(
x) $ may be found by functionally differentiating $H-\int dy \lambda
( y) C_{2}( y) $ with respect to $Q( x) $ to yield
\begin{align}
\label{XRef-Equation-104224150}%
Q_{\textup{mem}}( x) &=\frac{Q_{0}( x) \exp ( -\lambda ( y( x) )
) }{\int dx^{\prime }Q_{0}( x^{\prime }) \exp ( -\lambda ( y( x^{\prime
}) ) ) }
\\%
\label{XRef-Equation-104224138}%
 &\longrightarrow Q_{0}( x) f( y( x) )
\end{align}
where $\lambda ( y( x) ) $ is an undetermined Lagrange function
of $y( x) $. In Equation \ref{XRef-Equation-104224138} we present
a simpler notation by introducing an undetermined function $f( y(
x) ) $ to absorb the exponential function and the denominator term
that appeared in Equation \ref{XRef-Equation-104224150}. We may
impose the constraints in Equation \ref{XRef-Equation-10422420},
and use the definitions of $Q( y) $ and $P( y) $ in Equation \ref{XRef-Equation-104224212}
to obtain $f( y) $ in the form
\begin{equation}
f( y) =\frac{P( y) }{\int dx^{\prime }Q_{0}( x^{\prime }) \delta
( y-y( x^{\prime }) ) }
\end{equation}
and $Q_{\textup{mem}}( x) $ in the form
\begin{equation}
Q_{\textup{mem}}( x) =\frac{Q_{0}( x) P( y( x) ) }{\int dx^{\prime
}Q_{0}( x^{\prime }) \delta ( y( x) -y( x^{\prime }) ) }%
\label{XRef-Equation-104224422}
\end{equation}
Note that this result is a closed form solution because it contains
no undetermined Lagrange functions, unlike Equation \ref{XRef-Equation-104224355}
which contains an undetermined Lagrange vector $\lambda $. The normalisation
of this solution can be verified as follows
\begin{equation}
\begin{array}{rl}
 \int dx Q_{\textup{mem}}( x)  & =\int dx dy \delta ( y-y( x) )
\frac{Q_{0}( x) P( y) }{\int dx^{\prime }Q_{0}( x^{\prime }) \delta
( y-y( x^{\prime }) ) } \\
  & =\int dy P( y)  \\
  & =1
\end{array}
\end{equation}
where we use the identity $\int dy \delta ( y-y( x) ) =1$ to create
a dummy integral over $y$.
\subsection{Hierarchical maximum entropy method}\label{XRef-Section-104231312}

The purpose of this subsection is to present a generalisation of
Equation \ref{XRef-Equation-104224422} that uses hierarchical transformation
functions.

In practice the result in Equation \ref{XRef-Equation-104224422}
has a limited usefulness. Firstly, we would like to impose many
simultaneous constraints, each using its own constraint function
$\delta ( y_{i}-y_{i}( x) ) $ in Equation \ref{XRef-Equation-10422420},
but this cannot in general be done without sacrificing our closed
form solution in Equation \ref{XRef-Equation-104224422}. Secondly,
we would like to impose higher order constraints, using a constraint
function $\delta ( y-y( x) ) $. This may easily be done by making
the replacement $y\longrightarrow (y_{1},y_{2},\cdots )$ in Equation
\ref{XRef-Equation-104224422}. However, there is a hidden problem,
because the greater the dimensionality of $y$, the less easy is
it to make the necessary measurements to establish the form of $P(
y) $. Fortunately, there is a solution to both of these problems,
which we shall describe below.
\begin{figure}[h]
\begin{center}
\includegraphics[width=12cm]{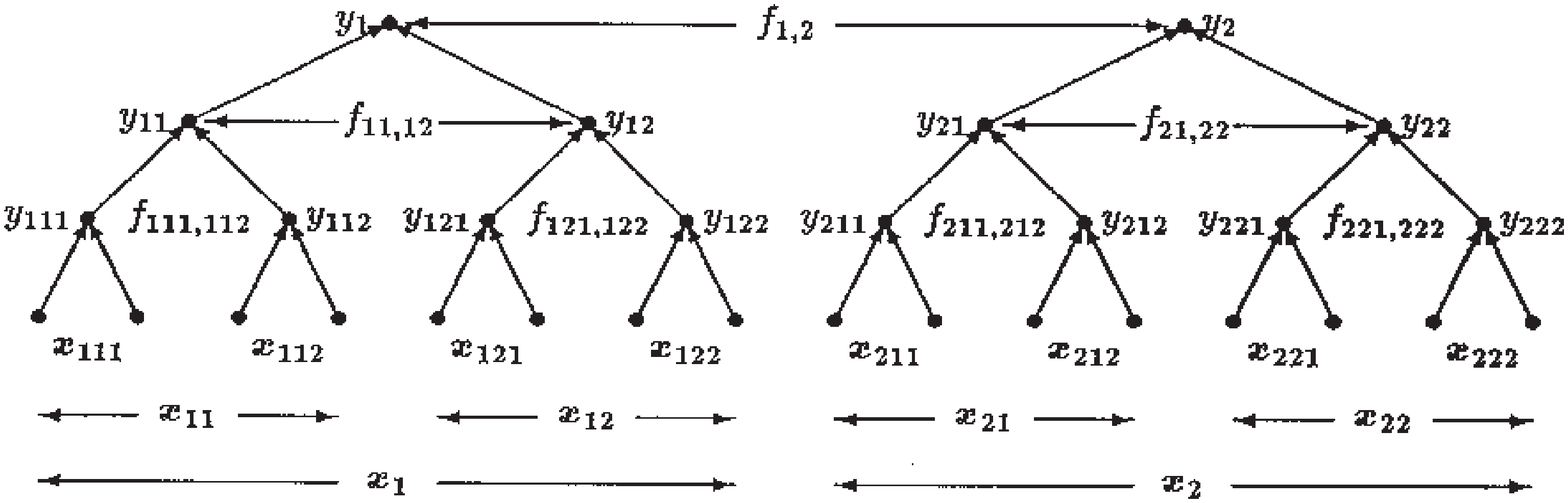}

\end{center}
\caption{Notation used in the hierarchical maximum entropy derivation.}\label{XRef-Figure-104224937}
\end{figure}

We shall apply the maximum entropy method with constraints of the
form shown in Equation \ref{XRef-Equation-10422420} to a hierarchy
of transformed versions of the input vector $x$. In order to make
our calculation tractable we introduce the notation shown in Figure
\ref{XRef-Figure-104224937}. The $x_{i j k ...}$ are various partitions
of the input vector $x$, the $y_{i j k ...}$ are various transformed
versions $y_{i j k ...}( x_{i j k ...}) $ of the input $x_{i j k
...}$, and the $f_{i j k \cdots ,i^{\prime }j^{\prime }k^{\prime
}\cdots }$ are the Lagrange functions $f_{i j k \cdots ,i^{\prime
}j^{\prime }k^{\prime }\cdots }( y_{i j k \cdots },y_{i^{\prime
}j^{\prime }k^{\prime }\cdots }) $ that appear in the generalised
version of the maximum entropy solution $Q_{\textup{mem}}( x) $
in\ \ Equation \ref{XRef-Equation-104224150}.

We choose to write the dependence of $y_{i j k ...}$ directly on
the input $x_{i j k ...}$, even though the value of $y_{i j k ...}$
is obtained via a number of intermediate transformations leading
from the leaf nodes of the tree up to node $i j k ...$, because
this leads to a transparent hierarchical maximum entropy derivation.
It is convenient to define $\Pi _{i j k \cdots }( x_{i j k \cdots
}) $ as the product of the Lagrange functions that appear beneath
node $i j k ...$ of the tree. $\Pi _{i j k \cdots }( x_{i j k \cdots
}) $ has the following recursion property
\begin{equation}
\begin{array}{rl}
 \Pi _{i j k \cdots }\left( x_{i j k \cdots }\right)  & =f_{i j
k \cdots  1,i j k \cdots  2}( y_{i j k \cdots  1}( x_{i j k \cdots
1}) ,y_{i j k \cdots  2}( x_{i j k \cdots  2}) )  \\
  & \begin{array}{cc}
   &
\end{array}\times \Pi _{i j k \cdots  1}( x_{i j k \cdots  1}) \Pi
_{i j k \cdots  2}( x_{i j k \cdots  2})
\end{array}%
\label{XRef-Equation-104225416}
\end{equation}
Also introduce a normalisation (or Jacobian) factor defined as
\begin{equation}
Z_{i j k \cdots }( y_{i j k \cdots }) =\int dx_{i j k \cdots } \delta
( y_{i j k \cdots }-y_{i j k \cdots }( x_{i j k \cdots }) ) \Pi
_{i j k \cdots }( x_{i j k \cdots }) %
\label{XRef-Equation-104225612}
\end{equation}
which is a sum of $\Pi _{i j k \cdots }( x_{i j k \cdots }) $ over
all the states $x_{i j k \cdots }$ of the leaf nodes beneath node
$i j k ...$ that are consistent with $y_{i j k ...}$ emerging at
node $i j k ...$.

The proof of the general hierarchical maximum entropy result proceeds
inductively. Firstly, we generalise Equation \ref{XRef-Equation-104223825}
to become
\begin{equation}
\begin{array}{rl}
 y_{i}( x)  & \longrightarrow \delta \left( y_{i j k \cdots  1}-y_{i
j k \cdots  1}( x_{i j k \cdots  1}) \right) \delta ( y_{i j k \cdots
2}-y_{i j k \cdots  2}( x_{i j k \cdots  2}) )  \\
 \lambda _{i} & \longrightarrow \lambda _{i j k \cdots }( y_{i j
k \cdots  1},y_{i j k \cdots  2})
\end{array}
\end{equation}
Secondly, we generalise Equation \ref{XRef-Equation-104224138} to
become
\begin{equation}
Q_{\textup{mem}}( x) =Q_{0}( x) f_{1,2}( y_{1}( x_{1}) ,y_{2}( x_{2})
) \Pi _{1}( x_{1}) \Pi _{2}( x_{2}) %
\label{XRef-Equation-104225320}
\end{equation}
where we display the Lagrange function $f_{1,2}( y_{1},y_{2}) $
that connects the topmost node-pair (i.e. node-pair $(1,2)$) in
the tree, but conceal the other Lagrange functions by using the
$\Pi _{i j k \cdots }$ notation.

We may determine the exact form of $f_{1,2}( y_{1},y_{2}) $ independently
of the rest of the Lagrange functions (which are hidden inside the
$\Pi _{1}( x_{1}) $ and $\Pi _{2}( x_{2}) $ functions) by imposing
the constraint shown in Equation \ref{XRef-Equation-10422420} and
Equation \ref{XRef-Equation-104224212} (as applied to node-pair
$(1,2)$) to obtain
\begin{equation}
\begin{array}{rl}
 P_{1,2}( y_{1},y_{2})  & =\int dx_{1}dx_{2}\delta ( y_{1}-y_{1}(
x_{1}) ) \delta ( y_{2}-y_{2}( x_{2}) ) Q_{\textup{mem}}( x)  \\
  & =f_{1,2}( y_{1},y_{2}) Z_{1}( y_{1}) Z_{2}( y_{2})
\end{array}
\end{equation}
which yields
\begin{equation}
f_{1,2}( y_{1},y_{2}) =\frac{P_{1,2}( y_{1},y_{2}) }{Z_{1}( y_{1})
Z_{2}( y_{2}) }
\end{equation}
Substituting this result back into Equation \ref{XRef-Equation-104225320}
yields
\begin{equation}
Q_{\textup{mem}}( x) =P_{1,2}( y_{1}( x_{1}) ,y_{2}( x_{2}) ) \frac{\Pi
_{1}( x_{1}) }{Z_{1}( y_{1}( x_{1}) ) }\frac{\Pi _{2}( x_{2}) }{Z_{2}(
y_{2}( x_{2}) ) }%
\label{XRef-Equation-104225529}
\end{equation}
which correctly obeys the constraint on the joint PDF $P_{1,2}(
y_{1},y_{2}) $ of the topmost pair of nodes in the tree.

We now marginalise $Q_{\textup{mem}}( x) $ in order to concentrate
our attention on the left-hand main branch of the tree. Thus
\begin{equation}
\begin{array}{rl}
 Q_{1,\textup{mem}}( x_{1})  & =\int dx_{2}Q_{\textup{mem}}( x)
\\
  & =\int dx_{2}dy_{2}\delta ( y_{2}-y_{2}( x_{2}) ) Q_{\textup{mem}}(
x)  \\
  & =P_{1}( y_{1}( x_{1}) ) \frac{\Pi _{1}( x_{1}) }{Z_{1}( y_{1}(
x_{1}) ) }
\end{array}%
\label{XRef-Equation-10423219}
\end{equation}
We now use the recursion property given in Equation \ref{XRef-Equation-104225416}
to extract the Lagrange function associated with node-pair $(11,12)$.
Thus $Q_{1,\textup{mem}}( x_{1}) $ becomes
\begin{equation}
Q_{1,\textup{mem}}( x_{1}) =P_{1}( y_{1}( x_{1}) ) f_{11,12}( y_{11}(
x_{11}) ,y_{12}( x_{12}) ) \frac{\Pi _{11}( x_{11}) \Pi _{12}( x_{12})
}{Z_{1}( y_{1}( x_{1}) ) }%
\label{XRef-Equation-104225455}
\end{equation}
As before, we may determine the exact form of $f_{11,12}( y_{11},y_{12})
$ independently of the rest of the Lagrange functions by applying
the constraints to node-pair $(11,12)$ to obtain
\begin{equation}
f_{11,12}( y_{11},y_{12}) =\frac{P_{11,12}( y_{11},y_{12}) }{P_{1}(
y_{1}) }\frac{Z_{1}( y_{1}) }{Z_{11}( y_{11}) Z_{12}( y_{12}) }%
\label{XRef-Equation-10422562}
\end{equation}
where the value of $y_{1}$ is to be understood to be obtained directly
from the values of $y_{11}$ and $y_{12}$ via the mapping which connects
node-pair $(11,12)$ to node 1. Substituting this result into Equation
\ref{XRef-Equation-104225455} yields
\begin{equation}
Q_{1,\textup{mem}}( x_{1}) =P_{11,12}( y_{11}( x_{11}) ,y_{12}(
x_{12}) ) \frac{\Pi _{11}( x_{11}) }{Z_{11}( y_{11}( x_{11}) ) }\frac{\Pi
_{12}( x_{12}) }{Z_{12}( y_{12}( x_{12}) ) }%
\label{XRef-Equation-104225534}
\end{equation}
By inspection, we see that Equation \ref{XRef-Equation-104225529}
and Equation \ref{XRef-Equation-104225534} are identical in form
once we have accounted for their different positions in the tree,
so we may use induction to obtain all of the rest of the Lagrange
functions in the form
\begin{equation}
\begin{array}{rl}
 f_{i j k \cdots  1,i j k \cdots  2}\left( y_{i j k \cdots  1},y_{i
j k \cdots  2}\right)  & =\frac{P_{i j k \cdots  1,i j k \cdots
2}( y_{i j k \cdots  1},y_{i j k \cdots  2}) }{P_{i j k \cdots }(
y_{i j k \cdots }) } \\
  & \begin{array}{cc}
   &
\end{array}\times \frac{Z_{i j k \cdots }( y_{i j k \cdots }) }{Z_{i
j k \cdots  1}( y_{i j k \cdots  1}) Z_{i j k \cdots  2}( y_{i j
k \cdots  2}) }
\end{array}%
\label{XRef-Equation-104225647}
\end{equation}
which is analogous to Equation \ref{XRef-Equation-10422562}, and
where $y_{\textup{ijk}\cdots }$ is obtained directly from the values
of $y_{i j k \cdots  1}$ and $y_{i j k \cdots  2}$. The $\frac{\Pi
}{Z}$ factors may be discarded once we reach the leaf nodes of the
tree, because the integral in Equation \ref{XRef-Equation-104225612}
then reduces to $Z=\Pi $.

Finally, by starting with Equation \ref{XRef-Equation-104225320}
and recursively simplifying the $\Pi _{\textup{ijk}\cdots }$ using
Equation \ref{XRef-Equation-104225416} and substituting for the
Lagrange functions $f_{i j k \cdots ,i^{\prime }j^{\prime }k^{\prime
}\cdots }$ using Equation \ref{XRef-Equation-104225647} we obtain
eventually for an $n$-layer tree
\begin{equation}
\begin{array}{rl}
 Q_{\textup{mem}}( x)  & =\left[ \prod \limits_{k=0}^{n-2}\prod
\limits_{i_{1},i_{2},\cdots ,i_{k}=1}^{2}\frac{P_{i_{1}i_{2} \cdots
i_{k}1,i_{1}i_{2} \cdots  i_{k}2}( y_{i_{1}i_{2} \cdots  i_{k}1}(
x_{i_{1}i_{2} \cdots  i_{k}1}) ,y_{i_{1}i_{2} \cdots  i_{k}2}( x_{i_{1}i_{2}
\cdots  i_{k}2}) ) }{P_{i_{1}i_{2} \cdots  i_{k}1}( y_{i_{1}i_{2}
\cdots  i_{k}1}( x_{i_{1}i_{2} \cdots  i_{k}1}) ) P_{i_{1}i_{2}
\cdots  i_{k}2}( y_{i_{1}i_{2} \cdots  i_{k}2}( x_{i_{1}i_{2} \cdots
i_{k}2}) ) }\right]  \\
  & \begin{array}{cc}
   &
\end{array}\times \left[ \prod \limits_{i_{1},i_{2},\cdots ,i_{k}=1}^{2}P_{i_{1}i_{2}
\cdots  i_{n}}( x_{i_{1}i_{2} \cdots  i_{n}}) \right]
\end{array}%
\label{XRef-Equation-104225740}
\end{equation}
where we have rearranged the terms to collect together the factors
that each node-pair $(i_{1}i_{2} \cdots  i_{k}1,i_{1}i_{2} \cdots
i_{k}2)$ contributes.

Although we have concentrated on deriving $Q_{\textup{mem}}( x)
$ for a binary tree, the principle of the derivation carries over
unchanged to arbitrary tree structures, and Equation \ref{XRef-Equation-104225740}
may easily be generalised. In Appendix B we explain the relationship
of the single layer version of Equation \ref{XRef-Equation-104225740}
to the random access memory network that is known as WISARD \cite{AleksanderStonham1979}.
\subsection{Diagrammatic notation}
\begin{figure}[h]
\begin{center}
\includegraphics[width=12cm]{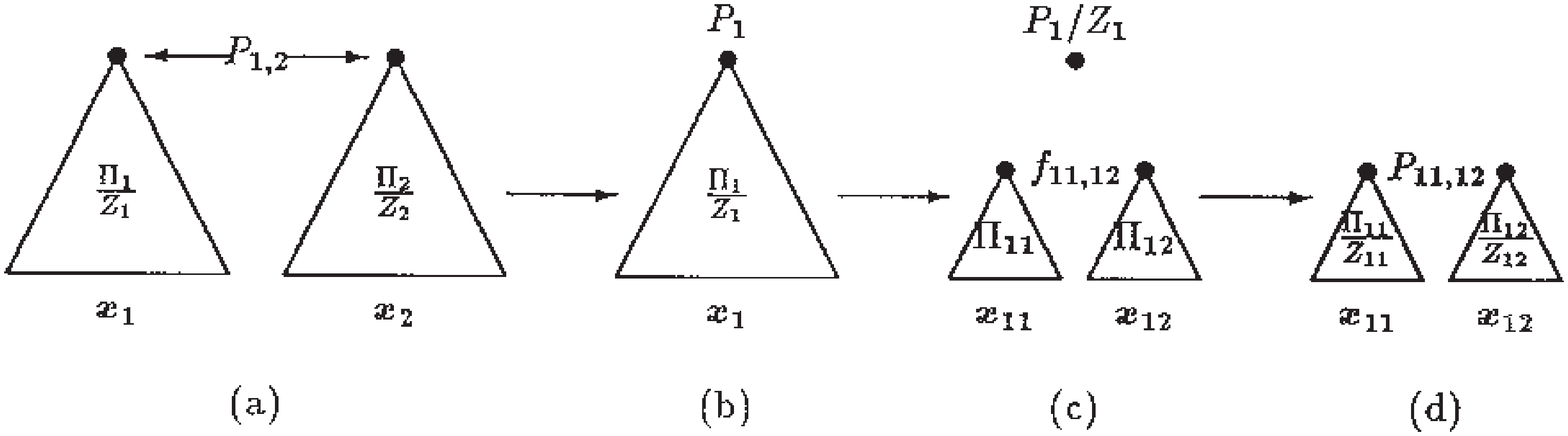}

\end{center}
\caption{The individual steps of the inductive hierarchical maximum
entropy derivation.}\label{XRef-Figure-10423036}
\end{figure}

We now present the steps in the inductive derivation leading from
Equation \ref{XRef-Equation-104225529} to Equation \ref{XRef-Equation-104225534}
as a diagram in Figure \ref{XRef-Figure-10423036}. We use a triangle
to represent a subtree, and we indicate its apex node, its associated
$\Pi $ or $\frac{\Pi }{Z}$ factor, and its dependence on $x$. Figure
\ref{XRef-Figure-10423036}a represents Equation \ref{XRef-Equation-104225529},
which is a pair of trees connected by the joint PDF of their apex
nodes. By integrating over $x_{2}$ we remove the right hand tree
to obtain Figure \ref{XRef-Figure-10423036}b, which corresponds
to Equation \ref{XRef-Equation-10423219}. We then explicitly display
the two daughter nodes to obtain Figure \ref{XRef-Figure-10423036}c,
which corresponds to Equation \ref{XRef-Equation-104225455}, although
we have grouped the terms together slightly differently, for simplicity.
This exposes one of the Lagrange functions which we determine explicitly
to obtain Figure \ref{XRef-Figure-10423036}d, which corresponds
to Equation \ref{XRef-Equation-104225534}. One cycle of the inductive
proof is completed by noting the correspondence between Figure \ref{XRef-Figure-10423036}a
and Figure \ref{XRef-Figure-10423036}d.
\begin{figure}[h]
\begin{center}
\includegraphics[width=9cm]{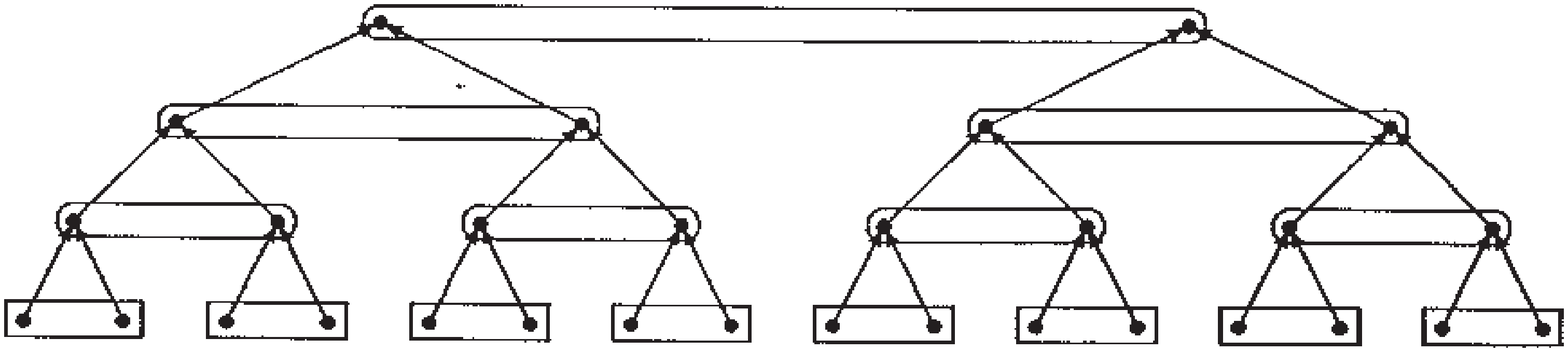}

\end{center}
\caption{A diagrammatic representation of the hierarchical maximum
entropy result.}\label{XRef-Figure-10423359}
\end{figure}

We represent Equation \ref{XRef-Equation-104225740} in diagrammatic
form in Figure \ref{XRef-Figure-10423359}. The tree structure represents
the flow of the transformations of the original input data $x$.
Each square cornered rectangle represents the marginal PDF of the
enclosed node-pair (i.e. one $P_{i_{1}i_{2} \cdots  i_{n}}$ term
from the second factor in Equation \ref{XRef-Equation-104225740}).
Each round cornered rectangle represents the normalised marginal
PDF of the encosed node-pair (i.e. one $\frac{P_{i_{1}i_{2} \cdots
i_{k}1,i_{1}i_{2} \cdots  i_{k}2}}{P_{i_{1}i_{2} \cdots  i_{k}1}P_{i_{1}i_{2}
\cdots  i_{k}2}}$ term from the first factor in Equation \ref{XRef-Equation-104225740}).
Overall, we obtain Equation \ref{XRef-Equation-104225740} as the
product of the rectangles in Figure \ref{XRef-Figure-10423359}.

This notation makes it easy to generalise the result in Equation
\ref{XRef-Equation-104225740} in a purely diagrammatic fashion,
by firstly constructing an arbitrary (i.e. not necessarily binary)
tree-like transformation of the input data, and secondly using as
maximum entropy constraints the marginal PDF of each set of sister
nodes in the tree. This prescription permits many possible ACE structures,
including those in which different constraints effectively operate
between different layers of the hierarchy (by mapping one or more
node values directly from layer to layer).

Each rectangle representing a marginal PDF in Figure \ref{XRef-Figure-10423359}
contributes to the maximum entropy estimate of the PDF of a cluster
of nodes in the input data. Because of the tree structure, clusters
at each length scale are built out of clusters at smaller length
scales. Equation \ref{XRef-Equation-104225740} tells us exactly
how to incorporate into $Q_{\textup{mem}}( x) $ any additional statistical
properties that might be observed when forming larger clusters out
of smaller clusters in this way.

Finally, Figure \ref{XRef-Figure-10423359} suggests an informal
derivation of Equation \ref{XRef-Equation-104225740}. Thus the expression
for the maximum entropy estimate of the joint PDF of the input data
$x$ in Equation \ref{XRef-Equation-104225529} can be viewed as the
joint PDF of the pair of nodes at the top of the tree in Figure
\ref{XRef-Figure-10423359} {\itshape times} corrective Jacobian
factors that compensate for effects of the many-to-one mapping that
the input data undergoes before it reaches the top of the tree.
The final maximum entropy expression in Equation \ref{XRef-Equation-104225740}
merely enumerates these corrective Jacobian factors explicitly in
terms of marginal PDFs measured at various levels of the tree. This
makes it clear that the maximum entropy method gives a result that
is consistent with simple counting arguments, which could therefore
be used in place of the rather involved maximum entropy derivation.
\section{Implementation of an anomaly detector}\label{XRef-Section-104222545}

Henceforth we shall refer to our hierarchical maximum entropy method
as an adaptive cluster expansion (ACE). In this section we describe
how to implement Equation \ref{XRef-Equation-104225740} in software.
We assume that the ACE transformation functions have already been
optimised using the unsupervised network training algorithm that
we describe in Appendix A and in \cite{Luttrell1989e}, so the purpose
of this section is to explain how to manipulate Equation \ref{XRef-Equation-104225740}
into a form that produces a useful output from the network. For
concreteness, we produce an output in the form of an image that
represents the degree to which each local patch of an input data
is statistically anomalous, when compared to the global statistical
properties of the input data.
\subsection{Two-dimensional array of inputs}
\begin{figure}[h]
\begin{center}
\includegraphics[width=4cm]{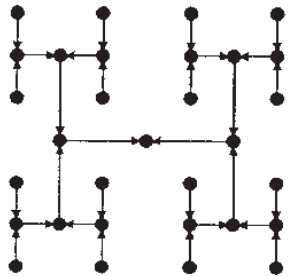}

\end{center}
\caption{ACE connectivity for processing a 2-dimensional array of
inputs.}\label{XRef-Figure-10423958}
\end{figure}

In Section \ref{XRef-Section-104222534} we represented ACE as if
it were operating on a 1-dimensional arrays of inputs (e.g. time
series). In practice this might indeed be the case, but in this
paper we choose to study 2-dimensional arrays of inputs (e.g. images).
There is no difficulty in applying ACE to an image, provided that
we appropriately assign the leaf nodes to pixels of the image. In
Figure \ref{XRef-Figure-10423958} we show the simplest possibility
in which the image is alternately compressed in the north-south
and east-west directions. A priori, the choice of whether to start
with\ \ north-south or east-west compression is arbitrary, but if
we knew, for instance, that the image had stronger short range correlations
in the east-west direction than the north-south direction, then
it would be better to compress east-west first of all. Note that
in Figure \ref{XRef-Figure-10423958} the topology of the tree is
the same as in Figure \ref{XRef-Figure-10423359}, but the way in
which the leaf nodes are identified with the data samples is different.

More generally, we could identify the leaf nodes of the tree with
the image pixels in any way that we please, provided that no pixel
is used more than once (to guarantee that the tree-like topology
is preserved). The problem of optimising the identification of leaf
nodes with pixels is extremely complicated, so we shall not pursue
it in this paper.
\subsection{Histograms}

The maximum entropy PDF in Equation \ref{XRef-Equation-104225740}
is a product of (normalised) marginal PDFs. In a practical implementation
of ACE the $y_{i j k \cdots }$ are discrete-valued quantities (for
instance, integers in the interval $[0,255]$), and the $P_{i j k
\cdots ,i^{\prime }j^{\prime }k^{\prime }\cdots }( y_{i j k \cdots
},y_{i^{\prime }j^{\prime }k^{\prime }\cdots }) $ are probabilities
(not PDFs). We estimate the $P_{i j k \cdots ,i^{\prime }j^{\prime
}k^{\prime }\cdots }( y_{i j k \cdots },y_{i^{\prime }j^{\prime
}k^{\prime }\cdots }) $ by constructing 2-dimensional histograms
\begin{equation}
P_{i j k \cdots ,i^{\prime }j^{\prime }k^{\prime }\cdots }( y_{i
j k \cdots },y_{i^{\prime }j^{\prime }k^{\prime }\cdots }) \simeq
\frac{1}{N_{i j k \cdots ,i^{\prime }j^{\prime }k^{\prime }\cdots
}}h_{i j k \cdots ,i^{\prime }j^{\prime }k^{\prime }\cdots }( y_{i
j k \cdots },y_{i^{\prime }j^{\prime }k^{\prime }\cdots }) %
\label{XRef-Equation-104231125}
\end{equation}
where $h_{i j k \cdots ,i^{\prime }j^{\prime }k^{\prime }\cdots
}( y_{i j k \cdots },y_{i^{\prime }j^{\prime }k^{\prime }\cdots
}) $ is the number of counts in the histogram bin $(y_{i j k \cdots
},y_{i^{\prime }j^{\prime }k^{\prime }\cdots })$, and $N$ is the
total number of histogram counts given by
\begin{equation}
N_{i j k \cdots ,i^{\prime }j^{\prime }k^{\prime }\cdots }=\sum
\limits_{y_{i j k \cdots }}\sum \limits_{y_{i^{\prime }j^{\prime
}k^{\prime }\cdots }}h_{i j k \cdots ,i^{\prime }j^{\prime }k^{\prime
}\cdots }( y_{i j k \cdots },y_{i^{\prime }j^{\prime }k^{\prime
}\cdots })
\end{equation}
Note that the estimate in Equation \ref{XRef-Equation-104231125}
suffers from Poisson noise due to the finite number of counts in
each histogram bin.

In order to build up this estimate we first of all train the ACE
transformation functions as explained in Appendix A. The histogram
bins are then initialised to zero, and subsequently filled with
counts by exposing the trained ACE to many examples of input vectors
(possibly, the set used to train the transformation functions).
Thus each vector is propagated up through the ACE-tree, and we then
inspect each node-pair $(i j k \cdots ,i^{\prime }j^{\prime }k^{\prime
}\cdots )$ for which a marginal probability needs to be estimated,
and increment its corresponding histogram bin thus
\begin{equation}
h_{i j k \cdots ,i^{\prime }j^{\prime }k^{\prime }\cdots }( y_{i
j k \cdots },y_{i^{\prime }j^{\prime }k^{\prime }\cdots }) \longrightarrow
h_{i j k \cdots ,i^{\prime }j^{\prime }k^{\prime }\cdots }( y_{i
j k \cdots },y_{i^{\prime }j^{\prime }k^{\prime }\cdots }) +1
\end{equation}
When the training set has been exhausted, histogram bin $(i j k
\cdots ,i^{\prime }j^{\prime }k^{\prime }\cdots )$ records the number
of times that state $(y_{i j k \cdots },y_{i^{\prime }j^{\prime
}k^{\prime }\cdots })$ occurred.

A major disadvantage of using histograms is that they have a large
number of adjustable parameters (i.e. the number of counts in each
bin) that have to be determined by the training data, so they do
not generalise very well. However, for the purpose of this paper,
we do not need to resort to using more sophisticated ways of estimating
PDFs.
\subsection{Translation invariant processing}

We wish to detect statistical anomalies in images which have otherwise
spatially homogeneous statistics, such as textures. An invariance
of the statistical properties of the true PDF $P( x) $ can be expressed
as
\begin{equation}
P( {\mathcal G} x) =P( x) %
\label{XRef-Equation-104231739}
\end{equation}
where ${\mathcal G}$ is any element of the invariance group, which
we shall assume to be the group of translations of the image pixels.
In Equation \ref{XRef-Equation-104225740} $Q_{\textup{mem}}( x)
$ does not respect translation invariance for two reasons. Firstly,
we use transformations $y_{i j k \cdots }( x_{i j k \cdots }) $
that are explicitly translation variant, because the functional
form depends on the $i j k \cdots $ indices. Secondly, we connect
together these transformations in translation variant way, because
the tree structure in Figure \ref{XRef-Figure-104224937} and Figure
\ref{XRef-Figure-10423958} does not treat all of its leaf nodes
equivalently. We shall therefore modify the cluster expansion procedure
that we derived in Section \ref{XRef-Section-104231312} to guarantee
translation invariance. This will lead to a much improved maximum
entropy estimate $Q_{\textup{mem}}( x) $ of the true $P( x) $.

Firstly, use the same transformation function at each position within
a single layer of ACE. Thus in Equation \ref{XRef-Equation-104225740}
we make the replacement
\begin{equation}
\begin{array}{rl}
 y_{i_{1}i_{2} \cdots  i_{k}1}( x_{i_{1}i_{2} \cdots  i_{k}1})
& \longrightarrow y^{k}( x_{i_{1}i_{2} \cdots  i_{k}1})  \\
 y_{i_{1}i_{2} \cdots  i_{k}2}( x_{i_{1}i_{2} \cdots  i_{k}2})
& \longrightarrow y^{k}( x_{i_{1}i_{2} \cdots  i_{k}2})
\end{array}%
\label{XRef-Equation-104231949}
\end{equation}
where we indicate that the transformation is associated with the
$k$-th layer of ACE by attaching a superscript $k$ to each function.
This yields
\begin{equation}
\begin{array}{rl}
 Q_{\textup{mem}}( x)  & =\left[ \prod \limits_{k=0}^{n-2}\prod
\limits_{i_{1},i_{2},\cdots ,i_{k}=1}^{2}\frac{P_{i_{1}i_{2} \cdots
i_{k}1,i_{1}i_{2} \cdots  i_{k}2}( y^{k}( x_{i_{1}i_{2} \cdots
i_{k}1}) ,y^{k}( x_{i_{1}i_{2} \cdots  i_{k}2}) ) }{P_{i_{1}i_{2}
\cdots  i_{k}1}( y^{k}( x_{i_{1}i_{2} \cdots  i_{k}1}) ) P_{i_{1}i_{2}
\cdots  i_{k}2}( y^{k}( x_{i_{1}i_{2} \cdots  i_{k}2}) ) }\right]
\\
  & \begin{array}{cc}
   &
\end{array}\times \left[ \prod \limits_{i_{1},i_{2},\cdots ,i_{k}=1}^{2}P_{i_{1}i_{2}
\cdots  i_{n}}( x_{i_{1}i_{2} \cdots  i_{n}}) \right]
\end{array}%
\label{XRef-Equation-104231452}
\end{equation}
Equation \ref{XRef-Equation-104231452} guarantees translation invariance
(in the sense of a ``single-instruction-multiple-data'' computer)
of the processing that occurs when the input data is propagated
upwards through the overlapping trees.

Secondly, assume that Equation \ref{XRef-Equation-104231739} holds
for all image translations, so that the marginal PDFs are independent
of position. We may make this explicit in our notation by making
the following replacement in Equation \ref{XRef-Equation-104231452}
\begin{equation}
\begin{array}{rl}
 \frac{P_{i_{1}i_{2} \cdots  i_{k}1,i_{1}i_{2} \cdots  i_{k}2}(
\cdot ) }{P_{i_{1}i_{2} \cdots  i_{k}1}( \cdot ) P_{i_{1}i_{2} \cdots
i_{k}2}( \cdot ) } & \longrightarrow \frac{P_{1,2}^{k}( \cdot )
}{P_{1}^{k}( \cdot ) P_{2}^{k}( \cdot ) } \\
 P_{i_{1}i_{2} \cdots  i_{n}}( \cdot )  & \longrightarrow P^{n-1}(
\cdot )
\end{array}%
\label{XRef-Equation-104231956}
\end{equation}
where we use the same superscript notation as in Equation \ref{XRef-Equation-104231949}.
This yields
\begin{equation}
\begin{array}{rl}
 Q_{\textup{mem}}( x)  & =\left[ \prod \limits_{k=0}^{n-2}\prod
\limits_{i_{1},i_{2},\cdots ,i_{k}=1}^{2}\frac{P_{1,2}^{k}( y^{k}(
x_{i_{1}i_{2} \cdots  i_{k}1}) ,y^{k}( x_{i_{1}i_{2} \cdots  i_{k}2})
) }{P_{1}^{k}( y^{k}( x_{i_{1}i_{2} \cdots  i_{k}1}) ) P_{2}^{k}(
y^{k}( x_{i_{1}i_{2} \cdots  i_{k}2}) ) }\right]  \\
  & \begin{array}{cc}
   &
\end{array}\times \left[ \prod \limits_{i_{1},i_{2},\cdots ,i_{k}=1}^{2}P^{n-1}(
x_{i_{1}i_{2} \cdots  i_{n}}) \right]
\end{array}%
\label{XRef-Equation-104231934}
\end{equation}
Equation \ref{XRef-Equation-104231934} guarantees not only translation
invariance of the transformations that propagate the data through
the tree, but also translation invariance of the marginal PDFs of
$P( x) $ that are used to construct $Q_{\textup{mem}}( x) $.

Both of the simplifications in Equation \ref{XRef-Equation-104231949}
and Equation \ref{XRef-Equation-104231956} reduce the total number
of unknowns that have to be determined. For a given amount of training
data we can thus construct a better maximum entropy estimate $Q_{\textup{mem}}(
x) $ of the true P(x). The transformation functions may be optimised
better, and the histogram bins have a reduced Poisson noise.

We usually apply ACE to such large input arrays that it is not appropriate
to build a single binary tree whose leaf nodes encompass the entire
input array. Instead, we divide the input array (which we shall
assume is a $2^{M}\times 2^{M}$ array of image pixels) into a set
of contiguous $2^{m_{1}}\times 2^{m_{2}}$ arrays, each of which
we analyse using Equation \ref{XRef-Equation-104231934}. There are
no constraint functions to measure the mutual dependencies between
these subarrays, so the maximum entropy joint PDF of the set of
subarrays is a product of terms of the form shown in Equation \ref{XRef-Equation-104231934}.
\begin{equation}
\begin{array}{rl}
 \log ( Q_{\textup{mem}}( x) )  & =\sum \limits_{k=0}^{n-2}\sum
\limits_{a_{1}=1}^{2^{M-m_{1}}}\sum \limits_{a_{2}=1}^{2^{M-m_{2}}}\sum
\limits_{i_{1},i_{2},\cdots ,i_{k}=1}^{2}\log ( \frac{P_{1,2}^{k}(
y^{k}( x_{i_{1}i_{2} \cdots  i_{k}1}^{a_{1},a_{2}}) ,y^{k}( x_{i_{1}i_{2}
\cdots  i_{k}2}^{a_{1},a_{2}}) ) }{P_{1}^{k}( y^{k}( x_{i_{1}i_{2}
\cdots  i_{k}1}^{a_{1},a_{2}}) ) P_{2}^{k}( y^{k}( x_{i_{1}i_{2}
\cdots  i_{k}2}^{a_{1},a_{2}}) ) })  \\
  & \begin{array}{cc}
   &
\end{array}+\log \left( \sum \limits_{a_{1}=1}^{2^{M-m_{1}}}\sum
\limits_{a_{2}=1}^{2^{M-m_{2}}}\sum \limits_{i_{1},i_{2},\cdots
,i_{k}=1}^{2}\log ( P^{n-1}( x_{i_{1}i_{2} \cdots  i_{n}}^{a_{1},a_{2}})
) \right)
\end{array}%
\label{XRef-Equation-10423217}
\end{equation}
The summation over $(a_{1},a_{2})$ ranges over the $2^{2M-m_{1}-m_{2}}$
contiguous subarrays in the overall $2^{M}\times 2^{M}$ array, and
the $a_{1},a_{2}$ superscript on each $x_{i j k \cdots }$ vector
indicates that it belongs to subarray $(a_{1},a_{2})$. Note that
we have transformed $Q_{\textup{mem}}( x) \longrightarrow \log (
Q_{\textup{mem}}( x) ) $ for convenience.

The final step in constructing a fully translation invariant PDF
is to modify the sum over subarrays so that it includes all possible
placements of the $2^{m_{1}}\times 2^{m_{2}}$ subarray within the
overall $2^{M}\times 2^{M}$ array. There are $2^{2M-m_{1}-m_{2}}$
possible positions when the placement of the subarray is restricted
as in Equation \ref{XRef-Equation-10423217}, whereas there are $(2^{M}-2^{m_{1}}+1)(2^{M}-2^{m_{2}}+1)$
possible positions when all placements of the subarray are permitted.
We therefore make the replacement
\begin{equation}
\begin{array}{rl}
 \sum \limits_{a_{1}=1}^{2^{M-m_{1}}}\sum \limits_{a_{2}=1}^{2^{M-m_{2}}}\left(
\cdot \right)  & \longrightarrow \frac{2^{2M-m_{1}-m_{2}}}{\left(
2^{M}-2^{m_{1}}+1\right) \left( 2^{M}-2^{m_{2}}+1\right) }\sum \limits_{p_{1}=1}^{2^{M}-2^{m_{1}}+1}\sum
\limits_{p_{2}=1}^{2^{M}-2^{m_{2}}+1}\left( \cdot \right)  \\
  & \simeq 2^{-m_{1}-m_{2}}\sum \limits_{p_{1}=1}^{2^{M}}\sum \limits_{p_{2}=1}^{2^{M}}\left(
\cdot \right)
\end{array}%
\label{XRef-Equation-104232215}
\end{equation}
in Equation \ref{XRef-Equation-10423217}, where $(p_{1},p_{2})$
is the coordinate of the pixel in the top left hand corner of the
$2^{m_{1}}\times 2^{m_{2}}$ subarray. If we ignore edge effects,
then we may use the approximation in the final line of Equation
\ref{XRef-Equation-104232215}, which is the average of $2^{m_{1}+m_{2}}$
separate contributions of the form shown in Equation \ref{XRef-Equation-10423217}.
Equation \ref{XRef-Equation-104232215} effectively replaces the
original maximum entropy PDF $Q_{\textup{mem}}( x) $ by the geometric
mean of a set of maximum entropy PDFs. This averaging reduces the
problems caused by Poisson noise on the histogram bin contents to
yield a greatly improved maximum entropy PDF estimate.
\begin{figure}[h]
\begin{center}
\includegraphics[width=12cm]{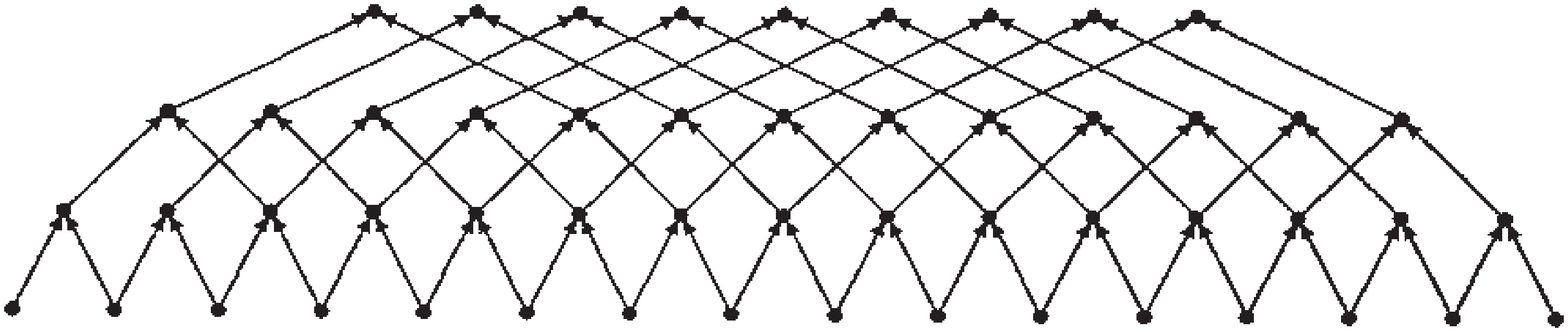}

\end{center}
\caption{Connectivity for multiple overlapping binary trees.}\label{XRef-Figure-104232412}
\end{figure}

In practice, we would implement each layer of ACE as a frame store,
and the transformation between each pair of adjacent layers as a
look-up table. The translation invariant ACE that we derived in
Equation \ref{XRef-Equation-10423217} (with the replacement given
in Equation \ref{XRef-Equation-104232215}) may be implemented using
the connectivity shown in Figure \ref{XRef-Figure-104232412}. Ignoring
edge effects, we may write Equation \ref{XRef-Equation-10423217}
symbolically as
\begin{equation}
\log ( Q_{\textup{mem}}( x) ) \simeq \sum \limits_{k=0}^{n-2}\frac{1}{2^{n-k}}\sum
\log ( \frac{P_{1,2}^{k}}{P_{1}^{k}P_{2}^{k}}) +\frac{1}{2}\sum
\log ( P^{n-1}) %
\label{XRef-Equation-10423268}
\end{equation}
where the inner summations range over all positions within a single
layer of Figure \ref{XRef-Figure-104232412}. We omit all of the
functional dependencies, because they are easy to obtain from Figure
\ref{XRef-Figure-10423359}. Each $\frac{P_{1,2}^{k}}{P_{1}^{k}P_{2}^{k}}$
term is represented by a rectangle with rounded corners in Figure
\ref{XRef-Figure-10423359}, and each $P^{n-1}$ term is represented
by a rectangle with square corners in Figure \ref{XRef-Figure-10423359}.
We have not drawn these rectangles in Figure \ref{XRef-Figure-104232412}
because they would overlap, and thus confuse the diagram.
\subsection{Forming a probability image}

Equation \ref{XRef-Equation-10423268} is the fundamental result
that we use to construct useful image processing schemes. However,
it would not be very useful simply to calculate the value of $\log
( Q_{\textup{mem}}) $ as a single global measure of the logarithmic
probability associated with an image. We choose instead to break
up Equation \ref{XRef-Equation-10423268} into smaller pieces, and
to examine their contribution to the overall $\log ( Q_{\textup{mem}})
$. In effect, we look at how $\log ( Q_{\textup{mem}}) $ is built
up from the information in each layer of ACE, which in turn we break
down into contributions from different areas of the image.
\begin{figure}[h]
\begin{center}
\includegraphics[width=9cm]{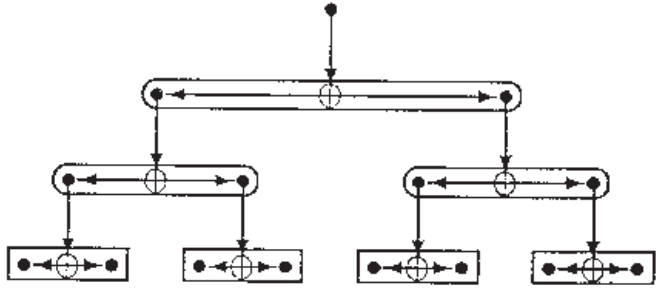}

\end{center}
\caption{Backpropagation scheme for constructing a probability image.}\label{XRef-Figure-10423275}
\end{figure}

In order to ensure that our decomposition of $\log ( Q_{\textup{mem}})
$ can be easily computed, we use the backpropagation scheme shown
in Figure \ref{XRef-Figure-10423275} to control the data flow through
a translation invariant network of an identical connectivity to
the one shown in Figure \ref{XRef-Figure-104232412}. Each node of
this backpropagation network records a logarithmic probability,
and is cleared to zero before starting the backpropagation computations.
The rectangles in Figure \ref{XRef-Figure-10423275} represent exactly
the same logarithmic probability terms that appeared in Figure \ref{XRef-Figure-10423359},
which we now use as sources of logarithmic probability that we inject
into the backpropagating data flow.

The detailed operation of Equation \ref{XRef-Equation-104231452}
is as follows. Each addition symbol takes as input a contribution
recorded at a node in the next layer above, adds its own logarithmic
probability source $\log ( \frac{P_{1,2}^{k}}{P_{1}^{k}P_{2}^{k}})
$, scales the result by $\frac{1}{4}$, and it finally adds a copy
of this result to the value stored at each of its own pair of associated
nodes, as shown. The values that accumulate at the leaf nodes represent
various contributions to the sum in Equation \ref{XRef-Equation-10423268}.
If the translation invariant version of Figure \ref{XRef-Figure-10423275}
is applied to the translation invariant network shown in Figure
\ref{XRef-Figure-104232412}, then the sum of the values that accumulate
at the leaf nodes reproduces Equation \ref{XRef-Equation-10423268}
precisely.

This method of computing $\log ( Q_{\textup{mem}}) $ might seem
to be circuitous, but it has the great advantage of both being computationally
cheap and forming an image-like representation of $\log ( Q_{\textup{mem}})
$, which we call a ``probability image''. Each $\log ( \frac{P_{1,2}^{k}}{P_{1}^{k}P_{2}^{k}})
$ term in Equation \ref{XRef-Equation-10423268} will contribute
equally to $2^{n-k}$ pixels in the probability image. These pixels
will be arranged as either a square or a 2-to-1 aspect ratio rectangle
according to whether there is an odd number or even number of backpropagation
steps from the {\itshape k}-th layer to the leaf nodes. The probability
image is therefore a superposition of square and rectangular tiles
of logarithmic probability. Each tile corresponds to a node of the
network shown in Figure \ref{XRef-Figure-104232412}.

It is useful to display as an image the contributions of a single
layer of the network to the probability image, because different
layers contribute to the structure of $\log ( Q_{\textup{mem}})
$ at different length scales. This image may be displayed in the
conventional way, with small probabilities mapped to black, large
probabilities mapped to white, and intervening probabilities mapped
to shades of grey, in which case we call it a ``probability image''.
It is also useful to invert the grey scale so that small probabilities
map to black, in which case we call it an ``anomaly image'', because
regions which have statistical properties that occur infrequently
show up as bright peaks in the image. We find that the use of probability
images and/or anomaly images is an extremely effective way of visually
interpreting $\log ( Q_{\textup{mem}}) $ in Equation \ref{XRef-Equation-10423268}.
\subsection{Modular implementation}
\begin{figure}[h]
\begin{center}
\includegraphics[width=12cm]{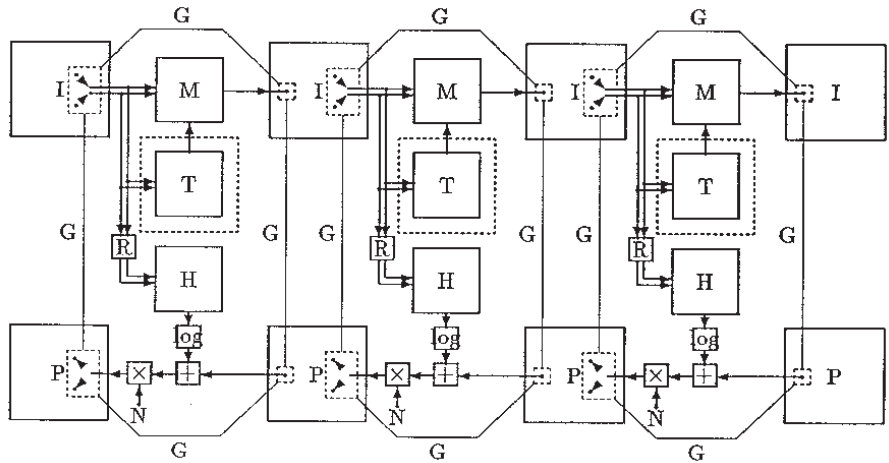}

\end{center}
\caption{Three layer translation invariant ACE system.}\label{XRef-Figure-104233212}
\end{figure}

For completeness we now present a brief description of a complete
system for producing probability and/or anomaly images. This system
consists of two tightly coupled subsystems - an ACE subsystem for
decomposing the image data, and a probability image subsystem for
forming the output image. Figure \ref{XRef-Figure-104233212} combines
in one diagram all of the results that we have discussed so far.
The upper part of Figure \ref{XRef-Figure-104233212} is a pure translation
invariant ACE subsystem, whereas the lower half is a backpropagating
probability image subsystem operating as shown in Figure \ref{XRef-Figure-10423275}.
The backpropagating subsystem takes input information from various
layers of ACE, as shown. Modules ``I'' are framestores that record
the various transformed images. Modules ``M'' are look-up tables
that record the inter-layer mappings. Modules ``T'' represent the
training algorithm that we explain in Appendix A, which we enclose
in a dashed box because the ``T'' modules are switched out of the
circuit once the mappings ``M'' have been determined. Modules ``H''
are accumulators that record the 2-dimensional histograms, and then
regularise and normalise them appropriately. Modules ``P'' are framestores
that record the various backpropagated probability images. Modules
``log'' are look-up tables (in fact only one such table is needed)
that implement a logarithm function. Modules ``$ \oplus\ \ $'' and
``$ \otimes\ \ $'' perform the addition and scaling operations that
we discussed earlier in connection with Figure \ref{XRef-Figure-10423275}.
``N'' is scaling factor (which is $\frac{1}{4}$ if we wish to reproduce
the result in Equation \ref{XRef-Equation-10423268}). The lines
that are annotated ``G'' represent a ganging together of the (pointers
to) pixels in adjacent layers of the ACE subsystem and in the probability
image subsystem. These ensure that the entire system works in lockstep,
as required.

The simplest mode of operation of this system can be broken down
into three stages Firstly, train each layer (from left to right)
of the ACE subsystem on a training image. Secondly, propagate a
test image (from left to right) through the layers of ACE. Finally,
construct a probability image by backpropagating (from right to
left) contributions from the various layers of ACE. Furthermore,
it is useful to display separately the probability (or anomaly)
images that emerge from each layer of ACE, as we shall see in Section
\ref{XRef-Section-104222552}.

There is a variety of methods of optimising ``T'', and hence ``M''.
The method that we describe in Appendix A trains each layer in sequence,
which takes 2.3 second per layer (using a VAXstation 3100, and assuming
6 bits per pixel), which gives a full training time of 20 seconds
for the 8 layer network that we use in our numerical simulations.
We do not make use of more sophisticated schemes in which different
layers are simultaneously trained, whilst communicating information
with each other to improve the global performance of ACE.
\subsection{Relationship to co-occurrence matrix methods}

Both the basic maximum entropy PDF $Q_{\textup{mem}}( x) $ in Equation
\ref{XRef-Equation-104225740}, and the translation invariant version
of $\log ( Q_{\textup{mem}}( x) ) $ in\ \ Equation \ref{XRef-Equation-10423268}
that we implement in practice, depend on various PDFs that are measured
in an ACE-tree. The second term of Equation \ref{XRef-Equation-10423268}
may be written as
\begin{equation}
Q_{\textup{mem}}( x) =\sqrt{\prod P^{n-1}}%
\label{XRef-Equation-104233649}
\end{equation}
Each $P^{n-1}$ factor is the spatial average of the marginal PDF
of pairs of adjacent pixel values, assuming that we use the identification
of leaf nodes with pixels that we show in Figure \ref{XRef-Figure-10423958}.
The square root in Equation \ref{XRef-Equation-104233649} compensates
for the fact that the product of $P^{n-1}$ factors generates the
product of two maximum entropy PDFs shifted by one pixel relative
to each other.

By using Equation \ref{XRef-Equation-104231125} we may approximate
Equation \ref{XRef-Equation-104233649} as a product of histograms.
In this case each histogram is the spatial average of the co-occurrence
matrix of pairs of adjacent pixel values, as commonly used in image
processing \cite{HaralickShanmugamDinstein1973}. Thus we may use
conventional co-occurrence matrix methods to construct a simple
form of maximum entropy PDF, which corresponds to using only one
layer of ACE.

This co-occurrence matrix result can be generalised, using Equation
\ref{XRef-Equation-104225740} or Equation \ref{XRef-Equation-10423268},
to model higher order statistical behaviour. Although these results
depend on co-occurrence matrices measured at various places in the
ACE-tree, the contributions which do not depend directly on the
input data (i.e. the first term of Equation \ref{XRef-Equation-10423268})
actually model higher order statistics of the input data. This is
because the value $y_{i j k \cdots }$ that emerges from node $i
j k \cdots $ of the ACE-tree depends on $x_{i j k \cdots }$, so
the joint PDF $P_{i j k \cdots  1,i j k \cdots  2}( y_{i j k \cdots
1},y_{i j k \cdots  2}) $ depends on the statistics of the pair
$(x_{i j k \cdots  1},x_{i j k \cdots  2})$. Thus ACE is a very
convenient way of combining together the various orders of statistical
information that are contained in co-occurrence matrices at various
places in the ACE-tree, as shown in Figure \ref{XRef-Figure-10423359}.
\section{Numerical results}\label{XRef-Section-104222552}

In this section we explain the finer details of how to implement
Figure \ref{XRef-Figure-104233212} in software, and we present the
results of applying the system to four $256\times 256$ images of
textiles taken from the Brodatz texture set \cite{Brodatz1966}.
\subsection{Experimental procedure}

We compensated for some of the effects of non-uniform illumination
by adding to each image a grey scale wedge whose gradient was chosen
in such a way as to remove the linear component of the non-uniformity.
Not only does this improve the translation invariance of the image
statistics, but it also improves the quality of the hierarchical
coding of the image, because we reduce the need to develop redundant
codes which differ only in their overall grey level.

Throughout our experiments we generate optimal inter-layer mappings
using the training methods that we explain in Appendix A. These
are known as topographic mappings in the neural network literature,
and we showed in \cite{Luttrell1989b} why they are appropriate for
building multistage vector quantisers. We choose to compress the
image in alternate directions using the following sequence: north/south,
east/west, north/south, east/west, etc. This compression sequence
leads to the following sequence of rectangular image regions that
influence the state of each pixel in each stage of ACE: $1\times
2$, $2\times 2$, $2\times 4$, $4\times 4$, etc, using (east/west,
north/south) coordinates. In all of our experiments we use an 8
stage ACE.

The number of bits per pixel that we use in each layer of ACE determines
the quality of the hierarchical vector quantisation that emerges.
Increasing the number of bits improves the quality of the vector
quantisation but increases the training time: we need to compromise
between these two conflicting requirements. In our work on simple
Brodatz texture images we have found that 6-8 bits per pixel is
sufficient.

It is important to note that for a given number of bits (after compression)
there is an upper limit on the allowed entropy that the input data
can have. This problem becomes more severe the greater the data
compression factor (i.e. the further we progress through the layers
of ACE). For instance, if the input image is very noisy then 6-8
bits will be sufficient only to give good vector quantisation performance
in the first few layers of ACE. This problem arises because ACE
does not have much prior knowledge of the statistical properties
of the input data, so each node of ACE encodes its input without
assuming a prior model. A prior model would allow us to reduce the
bit rate. This is a fundamental limitation to the capabilities of
the current version of ACE.

The choice of the size of the 2-dimensional histogram bins is also
important. A property of the topographic mappings that we use to
to connect the layers of ACE is that adjacent histogram bins derive
from input vectors that are close to each other (in the Euclidean
sense), so it is sensible to rebin the histogram by combining together
adjacent bins. Thus we control the histogram bin size by truncating
the low order bits of each binary vector that represents a pixel
value. If we do not truncate any bits, then the 2-dimensional histogram
faithfully records the number of times that a pair of pixel values
has occured. However, if we truncate $b$ low order bits of each
pixel value then effectively we sum together the histogram bins
in groups of $2^{2b}$ ($=2^{b}\times 2^{b}$) adjacent bins, which
smooths the histogram. The more smoothing that we impose the less
Poisson noise the histogram suffers. However, as we smooth the histogram
we run the danger of smoothing away significant structure that might
usefully be used to characterise the input image: so we need to
make a compromise. In our Brodatz texture work we use only 4-6 bits
of each pixel value to generate the histograms in each stage of
ACE. Note that we use more bits for vector quantisation than for
histogramming because the vector quantisation needs to be good enough
to preserve information for encoding by later layers of the hierarchy,
whereas the histogramming information is not passed to later layers.

In Equation \ref{XRef-Equation-10423268} we need to estimate the
logarithm of various probabilities from the histograms. We do this
in two stages. Firstly, we regularise the histograms by placing
a lower bound on the permitted number of counts. One possible prescription
is to ensure that each histogram bin has a number of counts at least
as large as the average number of counts in all the histogram bins
(as determined before regularising the histogram). Thus
\begin{equation}
h_{i j k \cdots ,i^{\prime }j^{\prime }k^{\prime }\cdots }( y_{i
j k \cdots },y_{i^{\prime }j^{\prime }k^{\prime }\cdots }) \longrightarrow
\left\{ \begin{array}{ccc}
 h_{i j k \cdots ,i^{\prime }j^{\prime }k^{\prime }\cdots }( y_{i
j k \cdots },y_{i^{\prime }j^{\prime }k^{\prime }\cdots })  &
& h\geq \left\langle  h\right\rangle   \\
 \left\langle  h_{i j k \cdots ,i^{\prime }j^{\prime }k^{\prime
}\cdots }( y_{i j k \cdots },y_{i^{\prime }j^{\prime }k^{\prime
}\cdots }) \right\rangle   &   & h<\left\langle  h\right\rangle
\end{array}\right. %
\label{XRef-Equation-109201552}
\end{equation}
where the angle brackets $\langle \cdots \rangle $ denote an average
over histogram bins, rounded up to the next largest integer to avoid
setting histogram bins to zero. Secondly, we estimate the probabilities
$P_{i j k \cdots ,i^{\prime }j^{\prime }k^{\prime }\cdots }( y_{i
j k \cdots },y_{i^{\prime }j^{\prime }k^{\prime }\cdots }) $ by
inserting the regularised histograms into Equation \ref{XRef-Equation-104231125}.
We use a marginalised version of Equation \ref{XRef-Equation-104231125}
to estimate the marginal probabilities $P_{i j k \cdots }( y_{i
j k \cdots }) $. Finally, we compute the logarithmic probabilities
in Equation \ref{XRef-Equation-10423268} by using a table of logarithms
of integers, up to the maximum possible number of counts that could
occur in a histogram bin - it suffices to tabulate logarithms up
to $\log ( N) $.

The prescription in Equation \ref{XRef-Equation-109201552} is crude
but effective. We could improve the performance by introducing prior
knowledge of the statistical properties of the input data. Our histogram
smoothing prescription already implicity makes use of prior knowledge
of the properties of the Posson noise process that affects the histogram
counts, and prior knowledge of the fact that adjacent histogram
bins correspond to similar input vectors. Additional prior knowledge
would further enhance the performance, especially in cases where
there is a limited amount of training data (such as small images,
or small segments of larger images).

A pitfall that must be avoided is using histogram bins that are
too small when one intends to train ACE on one image and then use
a different image to generate a probability image. Effectively,
the large number of small bins records the details of the statistical
fluctuations of the training image (as particular realisations of
a Poisson noise process in each bin), which thus acts as a detailed
record of the structure in the training image. The histograms thus
look very spiky, and in an extreme case there may be a counts recorded
in only a few bins with zeros in all of the remaining bins. If this
situation occurs then the training image records a large $\log (
Q_{\textup{mem}}( x) ) $, whereas a test image having the same statistical
properties records a small $\log ( Q_{\textup{mem}}( x) ) $. Effectively,
the spikes in the training and test image histograms are not coincident.
This problem can be solved by choosing a large enough histogram
bin size.

Finally, we display the logarithmic probability image as follows.
We determine the range of pixel values that occurs in the image,
and we translate and scale this into the range $[0,255]$. This ensures
that the smallest logarithmic probability appears as black, and
the largest logarithmic probability appears as white, and all other
values are linearly scaled onto intermediate levels of grey. This
prescription has its dangers because each probability image determines
its own special scaling, so one should be careful when comparing
two different probability images. It can also be adversely affected
by pixel value outliers arising from Poisson noise effects, where
an extreme value of a single pixel could affect the way in which
the whole of an image is displayed. However, we find that the overlapping
tree prescription in Figure \ref{XRef-Figure-104232412} together
with the backpropagation prescription in Figure \ref{XRef-Figure-10423275},
causes enough effective averaging together of the histogram bins
that we do not encounter problems with pixel value outliers.

In all of the images that we present below, we compensate for the
uneven illumination by introducing a grey scale wedge as we explained
earlier, we use 8 bits per pixel for vector quantisation, we use
6 bits per pixel for histogramming, and we invert the $[0,255]$
scale to produce an anomaly image, in which a white pixel indicates
a small (rather than a large) logarithmic probability.
\subsection{Texture 1}
\begin{figure}[h]
\begin{center}
\includegraphics{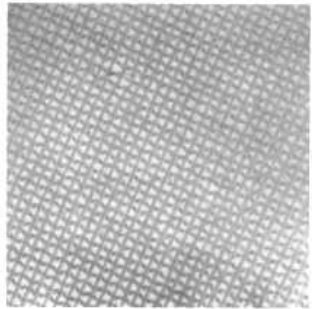}

\end{center}
\caption{$256\times 256$ image of Brodatz fabric number 1.}\label{XRef-Figure-109201910}
\end{figure}

In Figure \ref{XRef-Figure-109201910} we show the first Brodatz
texture image that we use in our experiments. The image is slightly
unevenly illuminated and has a fairly low contrast, but nevertheless
its statistical properties are almost translation invariant.
\begin{figure}[h]
\begin{center}
\includegraphics{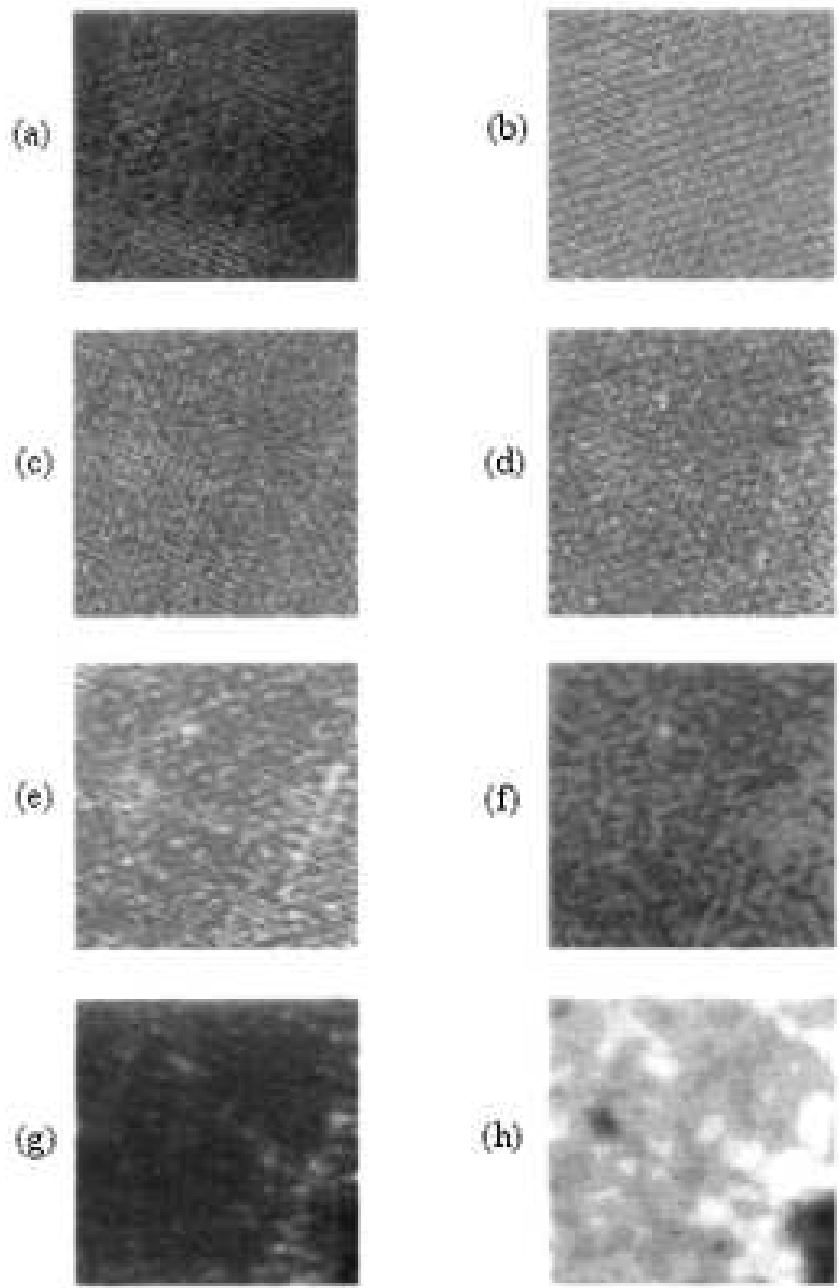}

\end{center}
\caption{$256\times 256$ anomaly images of Brodatz fabric number
1.}\label{XRef-Figure-109201957}
\end{figure}

In Figure \ref{XRef-Figure-109201957} we show the anomaly images
that derive from Figure \ref{XRef-Figure-109201910}. Note how the
anomaly images become smoother as we progress from Figure \ref{XRef-Figure-109201957}a
to Figure \ref{XRef-Figure-109201957}h, due to the increasing amount
of averaging that occurs amongst the overlapping backpropagated
rectangular tiles that build up each image.

Figure \ref{XRef-Figure-109201957}e and especially Figure \ref{XRef-Figure-109201957}f
reveal a highly localised anomaly in the original image. Figure
\ref{XRef-Figure-109201957}f corresponds to a length scale of $8\times
8$ pixels, which is the approximate size of the fault that is about
$\frac{1}{4}$ of the way down and slightly to the left of centre
of Figure \ref{XRef-Figure-109201910}. The fault does not show up
clearly on the other figures in Figure \ref{XRef-Figure-109201957}
because their characteristic length scales are either too short
or too long to be sensitive to the fault.

There is a major feature in the bottom right hand corner of Figure
\ref{XRef-Figure-109201957}h, where the anomaly image is darker
than average, indicating that the corresponding part of the original
image has a higher than average probability. This is a different
type of anomaly to the sort that we have envisaged so far - it occurs
because the corresponding part of original image happens to explore
only a high probability part of the space that is explored by the
whole image. This part of the anomaly image is surrounded by a brighter
than average border, which indicates a conventional anomalous region.

From Figure \ref{XRef-Figure-109201957} we conclude that ACE can
easily pick out localised faults in highly ordered textures.
\clearpage
\subsection{Texture 2}
\begin{figure}[h]
\begin{center}
\includegraphics{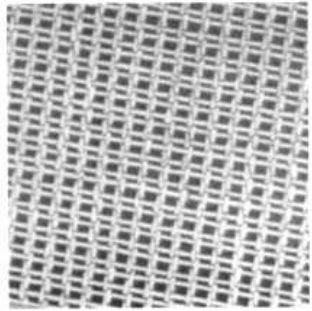}

\end{center}
\caption{$256\times 256$ image of Brodatz fabric number 2.}\label{XRef-Figure-109202317}
\end{figure}

In Figure \ref{XRef-Figure-109202317} we show the second Brodatz
texture image that we use in our experiments. The image has a high
contrast and translation invariant statistical properties.
\begin{figure}[h]
\begin{center}
\includegraphics{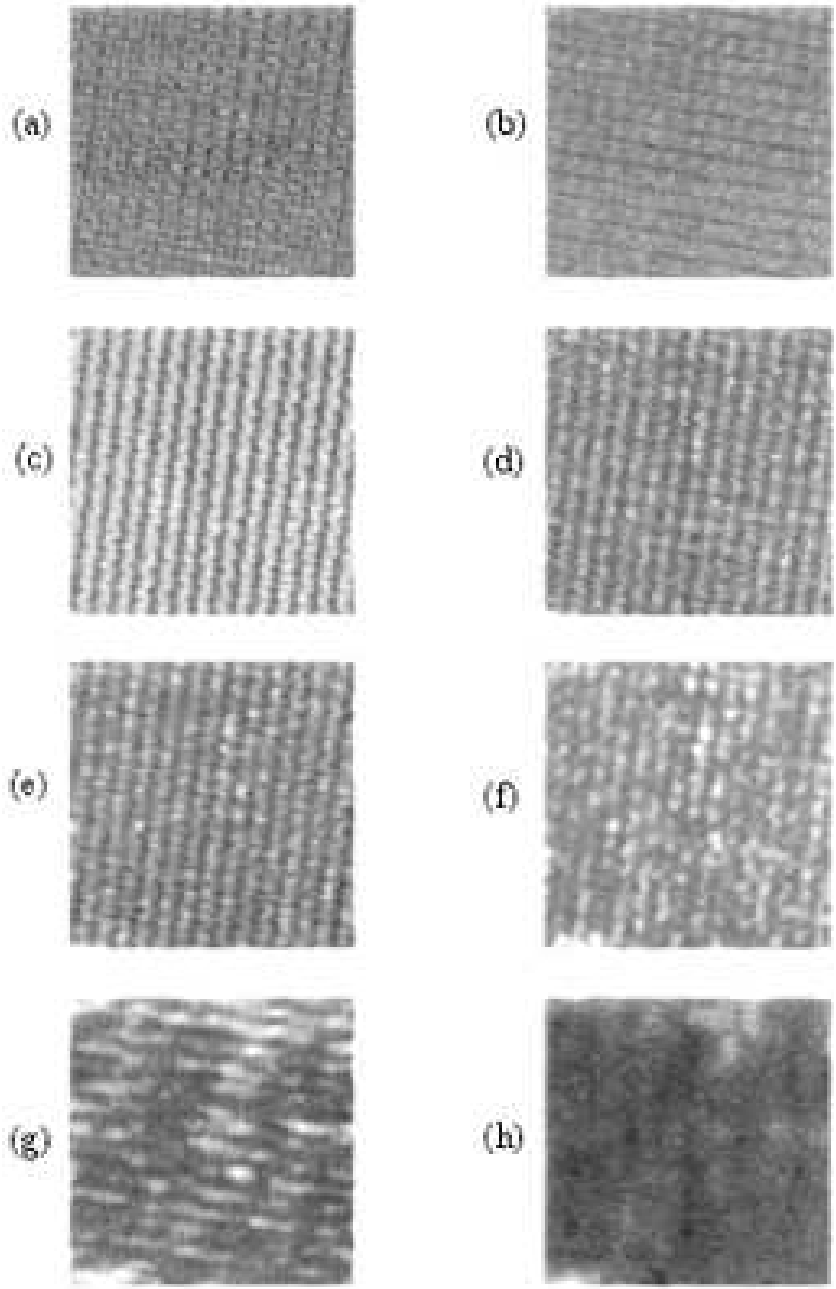}

\end{center}
\caption{$256\times 256$ anomaly images of Brodatz fabric number
2.}\label{XRef-Figure-10920243}
\end{figure}

In Figure \ref{XRef-Figure-10920243} we show the anomaly images
that derive from Figure \ref{XRef-Figure-109202317}. The most interesting
anomaly image is Figure \ref{XRef-Figure-10920243}f which shows
several localised anomalies. About halfway down and to the left
of centre of the image is an anomaly that corresponds to a dark
spot on the thread in Figure \ref{XRef-Figure-109202317}. The brightest
of the anomalies in the cluster just above the centre of the image
corresponds to what appears to be a slightly torn thread in Figure
\ref{XRef-Figure-109202317}. The other anomalies in this cluster
are weaker, and correspond to slight distortions of the threads.
There is another anomaly just below and to the right of the centre
of Figure \ref{XRef-Figure-10920243}g, which corresponds to what
appears to be another slightly torn thread in Figure \ref{XRef-Figure-109202317}.
These anomalies all occur at, or around, a length scale of $8\times
8$ pixels. Several of the anomaly images show an anomaly in the
bottom left hand corner of the image, which corrsponds to a small
uniform patch of fabric in Figure \ref{XRef-Figure-109202317}.

The results in Figure \ref{XRef-Figure-10920243} corroborate the
evidence in Figure \ref{XRef-Figure-109201957} that ACE can be trained
in an unsupervised fashion to pick out localised faults in highly
ordered textures.
\clearpage
\subsection{Texture 3}
\begin{figure}[h]
\begin{center}
\includegraphics{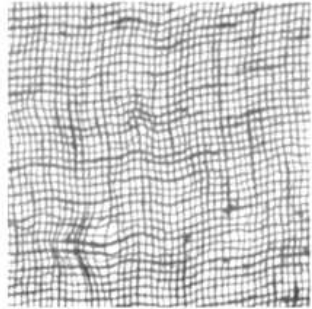}

\end{center}
\caption{$256\times 256$ image of Brodatz fabric number 3.}\label{XRef-Figure-109202710}
\end{figure}

In Figure \ref{XRef-Figure-109202710} we show the third Brodatz
texture image that use in our experiments.The image has a very high
contrast and statistical properties that are almost translation
invariant. However the density of anomalies is much higher than
in either Figure \ref{XRef-Figure-109201910} or Figure \ref{XRef-Figure-109202317}.
\begin{figure}[h]
\begin{center}
\includegraphics{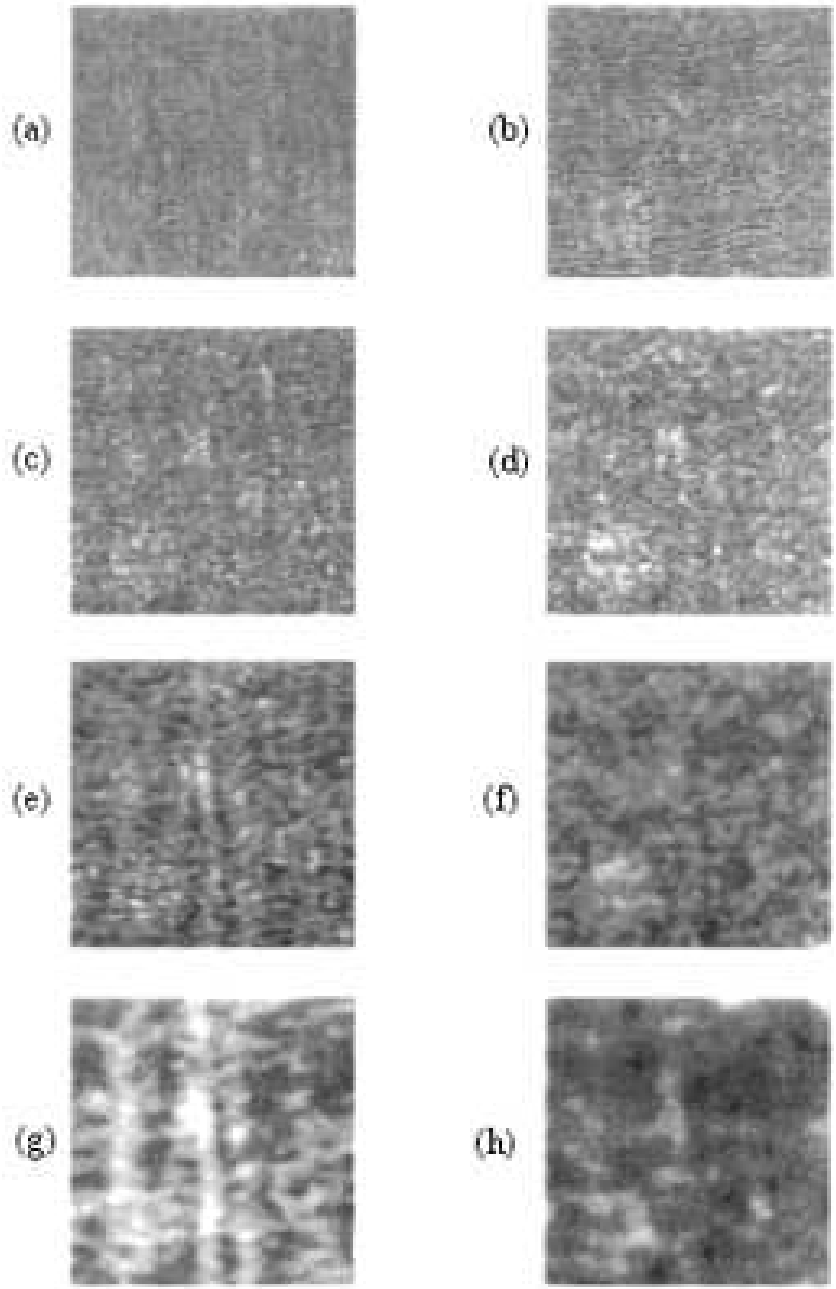}

\end{center}
\caption{$256\times 256$ anomaly images of Brodatz fabric number
3.}\label{XRef-Figure-109202845}
\end{figure}

In Figure \ref{XRef-Figure-109202845} we show the anomaly images
that derive from Figure \ref{XRef-Figure-109202710}. The most prominant
anomaly is in Figure \ref{XRef-Figure-109202845}g, at a length scale
of $8\times 16$ pixels, which corresponds to region of Figure \ref{XRef-Figure-109202710}
that is just above and to the left of centre of the image. This
region is anomalous because it is both distorted and has slightly
thicker threads than elsewhere. The large distorted region in the
bottom left hand corner of Figure \ref{XRef-Figure-109202710} does
not show up very clearly to the naked eye in Figure \ref{XRef-Figure-109202845},
but Figure \ref{XRef-Figure-109202845}f and Figure \ref{XRef-Figure-109202845}h
have significant peaks in this region. There are also many other
localised peaks in Figure \ref{XRef-Figure-109202845} which can
be traced back to corresponding faults in Figure \ref{XRef-Figure-109202710}.

Comparing Figure \ref{XRef-Figure-109202845} with Figure \ref{XRef-Figure-109201957}
and Figure \ref{XRef-Figure-10920243} we conclude that the ability
of ACE to pick out faults is degraded as the density of faults increases.
This is because the faults themselves are part of the statistical
properties that are extracted by ACE, and if a particular fault
occurs often enough in the image then it is no longer deemed to
be a fault.
\clearpage
\subsection{Texture 4}

In this section we present a slightly different type of experiment
in which we train ACE on one image and test ACE on another image.
To create the two images we start with a single $256\times 256$
image of a Brodatz texture, which we divide into a left half and
a right half. We then use the left half to build up the training
image, and the right half to build up the test image.
\begin{figure}[h]
\begin{center}
\includegraphics{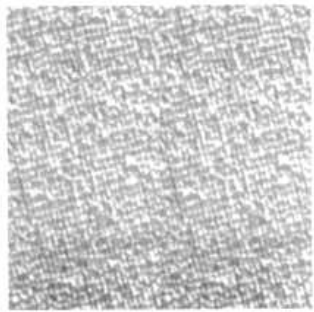}

\end{center}
\caption{$256\times 256$ image of Brodatz carpet for training.}\label{XRef-Figure-10920325}
\end{figure}
\begin{figure}[h]
\begin{center}
\includegraphics{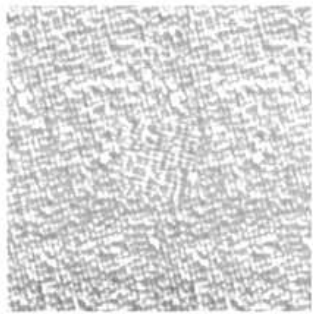}

\end{center}
\caption{$256\times 256$ image of Brodatz carpet for testing.}\label{XRef-Figure-109203218}
\end{figure}

In Figure \ref{XRef-Figure-10920325} we show the training image
which is a montage of two copies of the left hand half of a Brodatz
texture image. Note that this montage contains only as much information
as was present in the original half image from which it was constructed.
In Figure \ref{XRef-Figure-109203218} we show the test image which
is a montage of two copies of the right hand half of a Brodatz texture
image, and superimposed on that is a $64\times 64$ patch which we
generated by flipping the rows and columns of a copy of the top
left hand corner of this image. This patch is a hand crafted anomaly.
Note that in constructing these images we have scrupulously avoided
the possibility that the training and test images could contain
elements deriving from a common source.

\begin{figure}[h]
\begin{center}
\includegraphics{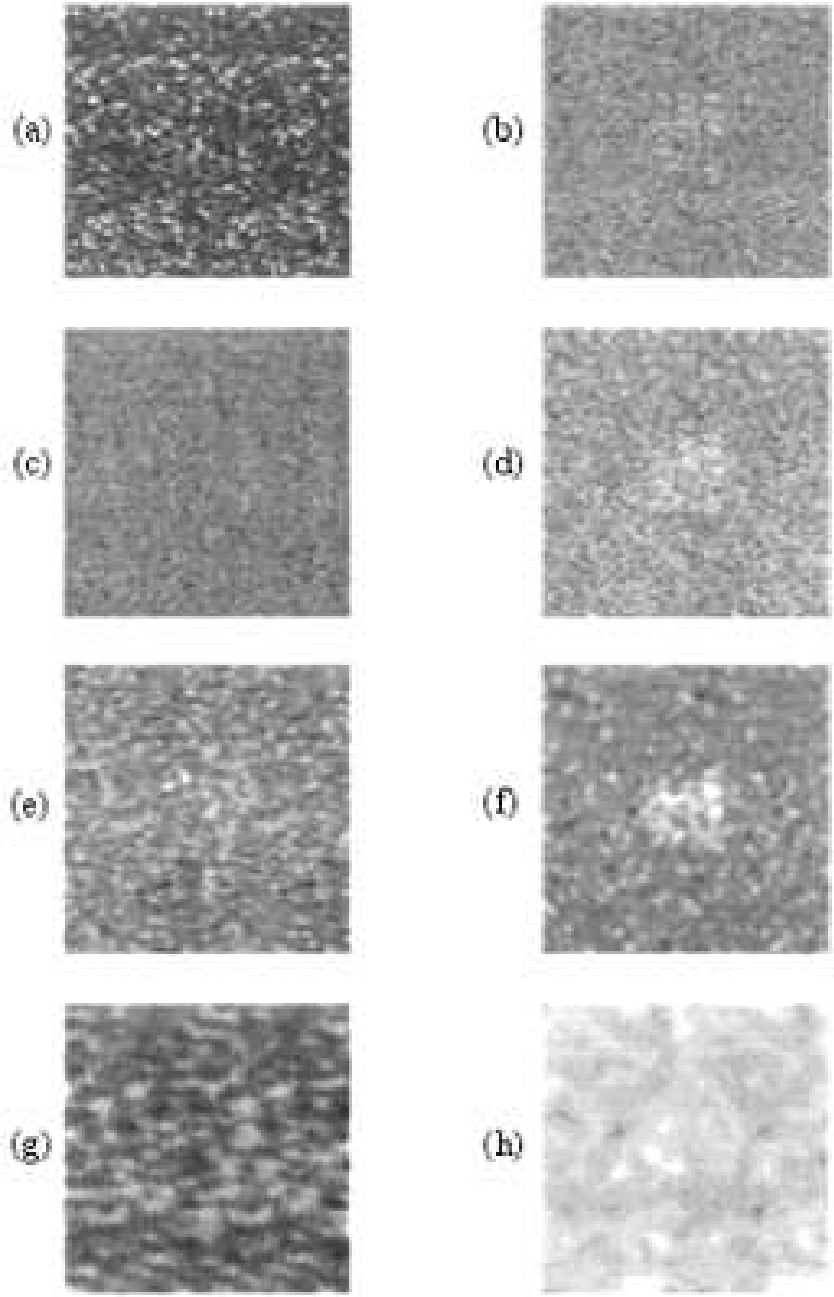}

\end{center}
\caption{$256\times 256$ anomaly images of Brodatz carpet.}\label{XRef-Figure-10920331}
\end{figure}
\clearpage
In Figure \ref{XRef-Figure-10920331} we show the anomaly images
that derive from Figure \ref{XRef-Figure-109203218} after having
trained on Figure \ref{XRef-Figure-10920325}. Figure \ref{XRef-Figure-10920331}f
shows the strongest response to the anomalous patch in the centre
of the image, corresponding to anomaly detection on a length scale
of $8\times 8$ pixels.
\section{Conclusions}

Using maximum entropy methods, we have shown how to construct maximum
entropy estimates of PDFs by using adaptive hierarchical transformation
functions to record various marginal PDFs of the data, which we
call an ``Adaptive Cluster Expansion'' (ACE). This method is a member
of the same family as the trainable MRF known as the Boltzmann Machine,
but it uses sophisticated transformations of the input data rather
than hidden variables to characterise the high order statistical
properties of the training set. The simulations in this paper use
hierarchical topographic mappings to build these transformations,
but this is a convenience, not a necessity.

We have also shown how to extend ACE so that it can be applied to
translation invariant image processing, such as the detection of
statistical anomalies in otherwise statistically homogeneous textures.
Our methods show great promise, not only because they are amenable
to a full theoretical analysis leading to closed-form maximum entropy
solutions, but also because they lead directly to a modular system
design which can locate anomalies in textures.

We have presented several examples where ACE successfully detects
anomalous regions in otherwise statistically homogeneous textures.
In all cases ACE adaptively extracts the global statistics of an
image at various length scales during the unsupervised training,
which takes 20 seconds (on a VAXStation 3100) for the 8 layer ACE
network that we applied to this problem. ACE then uses these statistics
to form an output image that represents the probability that each
local patch of the input image belongs to the ensemble of patches
presented during training. We call this a ``probability image''.

Some possible applications of our results are as follows. Inspection
of textiles: this relies on the assumed statistical homogeneity
of an unflawed piece of textile, so that faults show up as anomalies,
which we have demonstrated successfully in this paper. Detection
of targets in noisy background clutter in radar images: this is
basically a noisy version of the textile inspection problem, which
goes somewhat beyond what we have presented in this paper, because
it needs to address the problem of the noise entropy saturating
ACE. Texture segmentation: this is an ambitious goal which requires
much further analysis in order to derive a computationally cheap
method of handling multiple simultaneous textures.

\appendix\label{XRef-Appendix-104225752}

\section{Vector quantisation}

In this appendix we summarise the hierarchical vector quantisation
method that we presented in detail in \cite{Luttrell1989e}. In this
paper we use this technique to optimise the inter-layer mappings
in Figure \ref{XRef-Figure-104233212}. We have applied this technique
elsewhere to image compression \cite{Luttrell1989c}, and multilayer
self-organising neural networks \cite{{Luttrell1988a, Luttrell1989a,
Luttrell1989b, Luttrell1990b}}.
\subsection{Standard vector quantisation}\label{XRef-Section-109204826}

This subsection contains those details of the theory of standard
vector quantisation that one needs to understand before proceeding
to the modified vector quantisation scheme that we present in Section
\ref{XRef-Section-10920393}.

The problem is to form a coding $y$ of a vector $x$ in such a way
that a good estimate $x^{\prime }$ of $x$ can be constructed from
knowledge of $y$ alone. The sketch derivation in this section is
presented in greater detail in \cite{Luttrell1989b}. Thus a vector
quantiser is constructed by minimising a Euclidean distortion $D_{1}$
with respect to the choice of coding function $y( x) $ and decoding
function $x^{\prime }( y) $, where
\begin{equation}
D_{1}=\int dx P( x) \left\| x-x^{\prime }( y( x) ) \right\| ^{2}
\end{equation}
\begin{figure}[h]
\begin{center}
\includegraphics[width=9cm]{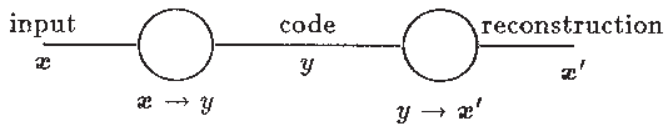}

\end{center}
\caption{Encoding and decoding in a vector quantiser.}\label{XRef-Figure-109204037}
\end{figure}

We may represent the encoding and decoding operations diagrammatically
as shown in Figure \ref{XRef-Figure-109204037}. By functionally
differentiating $D_{1}$ with respect to $y( x) $ and $x^{\prime
}( y) $ we obtain
\begin{align}
\label{XRef-Equation-109204156}%
\frac{\delta D_{1}}{\delta y( x) }&\left. =P( x)  \frac{\partial
}{\partial y}\left\| x^{\prime }( y) -x\right\| ^{2}\right| _{y=y(
x) }
\\%
\label{XRef-Equation-10920446}%
\frac{\delta D_{1}}{\delta x^{\prime }( y) }&=2\int dx P( x) \delta
( y-y( x) ) \left( x^{\prime }( y) -x\right)
\end{align}
Setting $\frac{\delta D_{1}}{\delta y( x) }=0$ in Equation \ref{XRef-Equation-109204156}
yields the optimum encoding function
\begin{equation}
y( x) =\arg  \begin{array}{c}
 \min  \\
 y
\end{array}\left\| x-x^{\prime }( y( x) ) \right\| ^{2}%
\label{XRef-Equation-109204633}
\end{equation}
which is called ``nearest neighbour encoding''. Setting $\frac{\delta
D_{1}}{\delta x^{\prime }( y) }=0$ in Equation \ref{XRef-Equation-10920446}
yields the optimum decoding function
\begin{equation}
x^{\prime }( y) =\frac{\int dx P( x) \delta ( y-y( x) ) x}{\int
dx P( x) \delta ( y-y( x) ) }%
\label{XRef-Equation-109204647}
\end{equation}
which is the update scheme derived in \cite{LindeBuzoGray1980}.
Alternatively, we may use an incremental scheme to optimise the
decoding function by following the path of steepest descent which
we may obtain from Equation \ref{XRef-Equation-10920446} as
\begin{equation}
\begin{array}{cc}
 \delta x^{\prime }( y) =\epsilon  \delta ( y-y( x) ) \left( x-x^{\prime
}( y) \right)  & \operatorname{where} 0<\epsilon <1.
\end{array}%
\label{XRef-Equation-109204654}
\end{equation}
An iterative optimisation scheme may be formed by alternately applying
Equation \ref{XRef-Equation-109204633} and then either Equation
\ref{XRef-Equation-109204647} or Equation \ref{XRef-Equation-109204654}.
This scheme will alternately improve the encoding and decoding functions
until a local minimum distortion is located. Alternating Equation
\ref{XRef-Equation-109204633} and Equation \ref{XRef-Equation-109204647}
is commonly called the ``LBG'' (after the authors of \cite{LindeBuzoGray1980})
or ``k-means'' algorithm.
\subsection{Noisy vector quantisation}\label{XRef-Section-10920393}

This subsection contains the theoretical details of the optimisation
of inter-layer mappings that we use in our numerical simulations
in Section \ref{XRef-Section-104222552}. Thus we generalise the
results of Section \ref{XRef-Section-109204826} to the case where
the encoded version of the input vector is distorted by a noise
process \cite{{KumazawaKasaharaNamekawa1984, Farvardin1990, Luttrell1989e,
Luttrell1990b}}.

Define a modified Euclidean distortion $D_{2}$ as
\begin{equation}
D_{2}=\int dx P( x) \int dn \pi ( n) \left\| x-x^{\prime }( y( x)
+n) \right\| ^{2}
\end{equation}
\begin{figure}[h]
\begin{center}
\includegraphics[width=12cm]{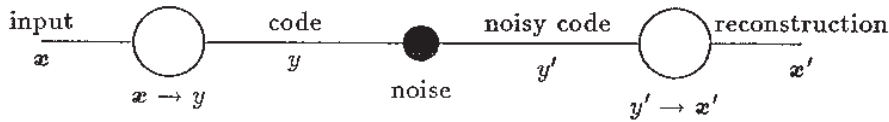}

\end{center}
\caption{Encoding and decoding in a noisy vector quantiser.}\label{XRef-Figure-109205127}
\end{figure}

We may represent the encoding and decoding operations together with
the noise process diagrammatically as shown in Figure \ref{XRef-Figure-109205127},
which is a trivially modified version of Figure \ref{XRef-Figure-109204037}.
By functionally differentiating $D_{2}$ with respect to $y( x) $
and $x^{\prime }( y) $ we obtain
\begin{align}
\label{XRef-Equation-109205235}%
\frac{\delta D_{2}}{\delta y( x) }&\left. =P( x) \int dn \pi ( n)
\frac{\partial }{\partial y}\left\| x^{\prime }( y) -x\right\| ^{2}\right|
_{y=y( x) +n}
\\%
\label{XRef-Equation-109205352}%
\frac{\delta D_{2}}{\delta x^{\prime }( y) }&=2\int dx P( x) \pi
( y-y( x) ) \left( x^{\prime }( y) -x\right)
\end{align}
Equation \ref{XRef-Equation-109205235} is a ``smeared'' version
of Equation \ref{XRef-Equation-109204156}, so $\frac{\delta D_{2}}{\delta
y( x) }=0$ does not lead to nearest neighbour encoding because the
distances to other code vectors have to be taken into account in
order to minimise the damaging effect of the noise process. However,
it is usually a good approximation to use the nearest neighbour
encoding scheme shown in Equation \ref{XRef-Equation-109204633}.
Setting $\frac{\delta D_{2}}{\delta x^{\prime }( y) }=0$ in Equation
\ref{XRef-Equation-109205352} yields the optimum decoding function
\begin{equation}
x^{\prime }( y) =\frac{\int dx P( x) \pi ( y-y( x) ) x}{\int dx
P( x) \pi ( y-y( x) ) }
\end{equation}
which should be compared with Equation \ref{XRef-Equation-109204647}.
Alternatively, we may obtain a steepest descent scheme in the form
\begin{equation}
\begin{array}{cc}
 \delta x^{\prime }( y) =\epsilon  \pi ( y-y( x) ) \left( x-x^{\prime
}( y) \right)  & \operatorname{where} 0<\epsilon <1
\end{array}%
\label{XRef-Equation-109205559}
\end{equation}
which should be compared with Equation \ref{XRef-Equation-109204654}.

As in Section \ref{XRef-Section-109204826}, iterative optimisation
schemes can be constructed in which we alternate the optimisation
of the coding and decoding functions. Alternating Equation \ref{XRef-Equation-109204633}
(which approximately solves $\frac{\delta D_{2}}{\delta x^{\prime
}( y) }=0$) and Equation \ref{XRef-Equation-109205559} yields the
standard topographic mapping training algorithm \cite{Kohonen1984},
which is widely used in various forms in neural network simulations.
\subsection{Hierarchical vector quantisation}

\begin{figure}[h]
\begin{center}
\includegraphics[width=12cm]{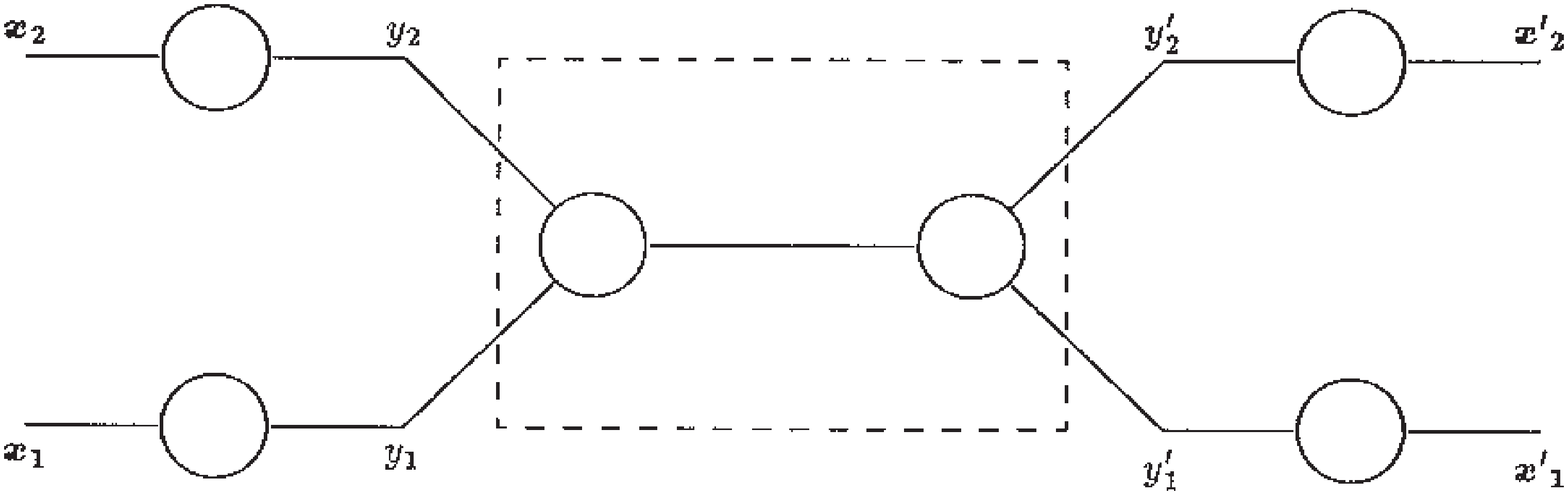}

\end{center}
\caption{Encoding and decoding in a hierarchical vector quantiser.}\label{XRef-Figure-109205734}
\end{figure}

In Figure \ref{XRef-Figure-109205734} we show the simplest type
of hierarchical vector quantiser. It consists of an inner quantiser
contained in the dashed box, surrounded by a pair of outer quantisers.
If the part of the diagram contained in the dashed box were removed
and direct connections made so that $y_{1}^{\prime }=y_{1}$ and
$y_{2}^{\prime }=y_{2}$, then Figure \ref{XRef-Figure-109205734}
would reduce to a pair of independent vector quantisers of the type
shown in Figure \ref{XRef-Figure-109204037}. The dashed box contains
a vector quantiser which encodes $(y_{1},y_{2})$ to produce a code
which it subsequently decodes to obtain $(y_{1}^{\prime },y_{2}^{\prime
})$.

From the point of view of $y_{1}$ the effect of being passed through
the inner quantiser is to modify $y_{1}$ thus $y_{1}\longrightarrow
y_{1}^{\prime }$. A similar argument applies to $y_{2}\longrightarrow
y_{2}^{\prime }$. The actual distortions $y_{1}^{\prime }-y_{1}$
and $y_{2}^{\prime }-y_{2}$ will be correlated in practice, but
we shall model them as if they were independent processes, and thus
reduce Figure \ref{XRef-Figure-109205734} to two independent vector
quantisers of the type shown in Figure \ref{XRef-Figure-109205127}.

This procedure can be extended to a hierarchical vector quantiser
with any number of levels of nesting. From the point of view of
the quantisers at any level, we shall model the effect of the quantisers
inwards from that level as independent distortion processes. It
turns out not to be critically important what precise distortion
model one uses, provided that it approximately represents the overall
scale of the distortion due to quantisation.

In \cite{Luttrell1989e} we presented in detail a phenomenological
distortion model that we used to obtain an efficient training procedure
for topographic mappings and their application to hierarchical vector
quantisers. Alternatively, the standard topographic mapping training
procedure in \cite{Kohonen1984} could be used, but this is a rather
inefficient algorithm. The basic training procedure may be obtained
from Equation \ref{XRef-Equation-109205559} as
\begin{enumerate}
\item Select a training vector $x$ at random from the training set.\label{XRef-Item1Numbered-1092115}
\item Encode $x$ to produce $y$ ($=y( x) $).
\item For all $y^{\prime }$ do the following:
\begin{enumerate}
\item Determine the corresponding code vector $x^{\prime }( y^{\prime
}) $.
\item Move the code vector\ \ $x^{\prime }( y^{\prime }) $ directly
towards the input vector $x$ by a distance $\epsilon  \pi ( y-y(
x) ) |x-x^{\prime }( y) |$.
\end{enumerate}
\item Go to step \ref{XRef-Item1Numbered-1092115}.
\end{enumerate}
This cycle is repeated as often as is required to ensure convergence
of the codebook of code vectors.

The standard method \cite{Kohonen1984} specifies that $\pi ( y^{\prime
}-y) $ should be an even unimodal function whose width should be
gradually decreased as training progresses. This allows coarse-grained
organisation of the codebook to occur, followed progressively by
ever more fine-grained organisation, until finally the algorithm
converges towards an optimum codebook.

In our own modification \cite{Luttrell1989e} of the standard method
we replace a shrinking $\pi ( y^{\prime }-y) $ function acting on
a fixed number of code vectors by a fixed $\pi ( y^{\prime }-y)
$ function acting on an increasing number of code vectors. There
are many minor variations on this theme, but we find that it is
sufficient to define
\begin{equation}
\pi ( y^{\prime }-y) =\left\{ \begin{array}{ccc}
 \epsilon  &   & y^{\prime }=y \\
 \epsilon ^{\prime } &   & \left| y^{\prime }-y\right| =1 \\
 0 &   & \left| y^{\prime }-y\right| >1
\end{array}\right.
\end{equation}
where we have absorbed $\epsilon $ in Equation \ref{XRef-Equation-109205559}
into the definition of $\pi ( y^{\prime }-y) $. We use a binary
sequence of codebook sizes $N=2,4,8,16,32,\cdots $, where each codebook
is initialised by interpolation from the next smaller codebook.
We find that the following parameter values yield adequate convergence:
$\epsilon =0.1$, $\epsilon ^{\prime }=0.05$, and we perform $20N$
training updates before doubling the value of $N$ and progressing
to the next larger size of codebook. The $N=2$ codebook can be initialised
using a random pair of vectors from the training set.

\section{Relationship to WISARD}
\begin{figure}[h]
\begin{center}
\includegraphics[width=12cm]{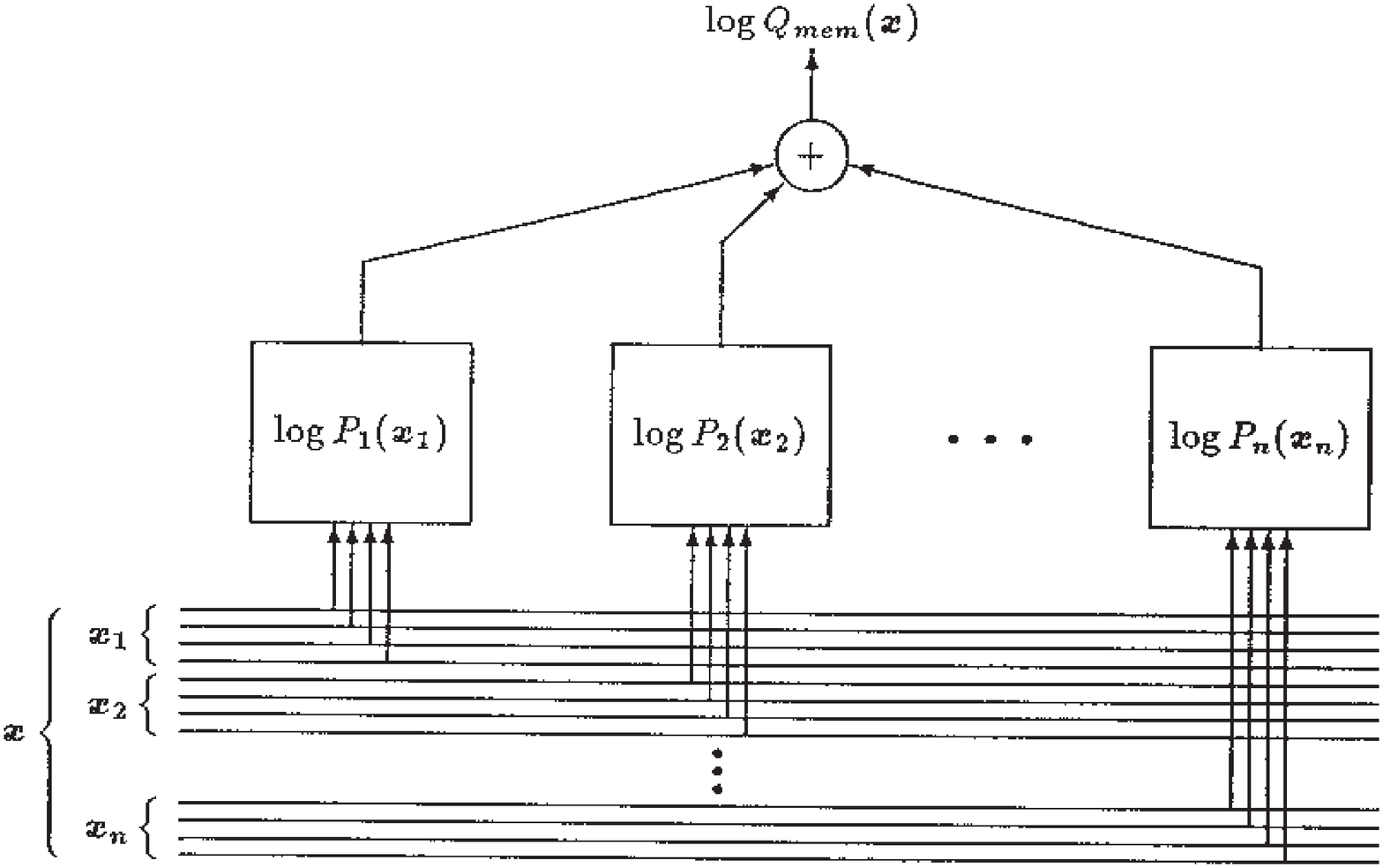}

\end{center}
\caption{Single layer ACE is WISARD.}\label{XRef-Figure-10921335}
\end{figure}

The second bracketed term in Equation \ref{XRef-Equation-104225740}
could be implemented in hardware as shown in Figure \ref{XRef-Figure-10921335}.
This implementation assumes that the state vector $x$ is quantised
by representing each of its components using a finite number of
binary digits (bits). Note that we have taken advantage of the fact
that we are discussing a single layer network in order to simplify
the notation in Figure \ref{XRef-Figure-10921335} (as compared with
Equation \ref{XRef-Equation-104225740}). The $i$-th block in this
circuit is a random access memory (RAM) which records a transformation
from $x_{i}$ (the address of an entry in the RAM) to $\log  P_{i}(
x_{i}) $ (the corresponding entry in the RAM). The data bus at the
bottom of Figure \ref{XRef-Figure-10921335} carries the components
of $x$ (represented bitwise) to the relevant RAM. Note that each
bit of $x$ is used exactly once in forming addresses for the RAM,
so the mapping from $x$ to the set of addresses is {\itshape bijective}.
The upper part of Figure \ref{XRef-Figure-10921335} shows how the
outputs are directed to an accumulator where they are summed to
form $\log  Q_{\textup{mem}}( x) $.

Figure \ref{XRef-Figure-10921335} is a variant of the WISARD pattern
recognition network \cite{AleksanderStonham1979}. The elements that
our MEM solution and WISARD have in common are: a bijective mapping
from the bits of an input state vector onto the address lines of
a set of RAMs, and the accumulation of the outputs of the RAMs to
form the overall network output.

However, there are some differences between the single layer ACE
and the WISARD prescriptions for the contents of the RAMs. ACE specifies
a set of functions (i.e. logarithms of marginal probabilities) to
tabulate in the RAMs. Suppose that we truncate these table entries
to a 1-bit representation, so that we use 0 to represent small logarithmic
probabilities and 1 to represent large logarithmic probabilities.
Each entry (i.e. 1 or 0) in the table then records whether the configuration
of binary digits (i.e. the address of the entry) frequently occurs
in the set of patterns corresponding to $P( x) $ (the ``training
set''). The final output is therefore the total number of 1's that
the input pattern addresses in the $n$ tables. In effect, this is
the total number of coincidences between configurations of bits
in the input pattern and those in a predefined category. This 1-bit
version of ACE is qualitatively the same as the table look-up and
summation operations performed in the simplest WISARD network, which
completes the connection that we sought between ACE and WISARD.\label{TitleNote}


\begin{thebibliography}{000}
\bibitem{Luttrell1987b} Luttrell, S. P. (1987). \textit{Markov random
fields: a strategy for clutter modelling}. In \textit{Proc. AGARD
Conf. on Scattering and Propagation in Random Media} (pp. 7.1--7.8).\label{Luttrell1987b}
\bibitem{Luttrell1987c} Luttrell, S. P. (1987). \textit{The use
of Markov random field models in sampling scheme design}. In \textit{Proc.
SPIE Int. Symp. on Inverse Problems in Optics} (pp. 182--188). New
York: SPIE.\label{Luttrell1987c}
\bibitem{Luttrell1987d} Luttrell, S. P. (1987). \textit{Designing
Markov random field structures for clutter modelling}. In \textit{Radar
87} (pp. 222--226). London: IEE.\label{Luttrell1987d}
\bibitem{Luttrell1987a} Luttrell, S. P. (1987). \textit{The use
of Markov random field models to derive sampling schemes for inverse
texture problems}. \textit{Inverse Problems}, \textbf{3}(2), 289--300.\label{Luttrell1987a}
\bibitem{Luttrell1988c} Luttrell, S. P. (1988). \textit{A maximum
entropy approach to sampling function design}. \textit{Inverse Probl.},
\textbf{4}(3), 829--841.\label{Luttrell1988c}
\bibitem{AckleyHintonSejnowski1985} Ackley, D. H., Hinton, G. E.,
\& Sejnowski, T. J. (1985). \textit{A learning algorithm for Boltzmann
machines}. \textit{Cognitive Sci.}, \textbf{9}, 147--169. \label{AckleyHintonSejnowski1985}
\bibitem{HaralickShanmugamDinstein1973} Haralick, R. M., Shanmugam,
K., \& Dinstein, I. (1973). \textit{Textural features for image
classification}. \textit{IEEE Trans. Syst. Man Cyb.}, \textbf{3}(6),
610--621.\label{HaralickShanmugamDinstein1973}
\bibitem{Luttrell1989d} Luttrell, S. P. (1989). \textit{The use
of Bayesian and entropic methods in neural network theory}. In G.
Erickson, J. T. Rychert  \& C. R. Smith  (Ed.), \textit{Maximum
Entropy and Bayesian Methods} (pp. 363--370). Dordtrecht: Kluwer.\label{Luttrell1989d}
\bibitem{Kohonen1984} Kohonen, T. (1984). \textit{Self-Organisation
and Associative Memory}. Berlin: Springer-Verlag.\label{Kohonen1984}
\bibitem{Luttrell1988a} Luttrell, S. P. (1988). \textit{Self organising
multilayer topographic mappings}. In \textit{Proc. IEEE Conf. on
Neural Networks} (pp. 93--100).\label{Luttrell1988a}
\bibitem{Luttrell1988b} Luttrell, S. P. (1988). \textit{Image compression
using a neural network}. In \textit{Proc. IGARSS Conf. on Remote
Sensing} (pp. 1231--1238).\label{Luttrell1988b}
\bibitem{Luttrell1989a} Luttrell, S. P. (1989). \textit{Self-organisation:
a derivation from first principles of a class of learning algorithms}.
In \textit{Proc. IEEE Conf. on Neural Networks} (pp. 495--498).\label{Luttrell1989a}
\bibitem{Luttrell1989e} Luttrell, S. P. (1989). \textit{Hierarchical
vector quantisation}. \textit{Proc. Inst. Electr. Eng. I}, \textbf{136}(6),
405--413.\label{Luttrell1989e}
\bibitem{Luttrell1990b} Luttrell, S. P. (1990). \textit{Derivation
of a class of training algorithms}. \textit{IEEE Trans. Neural Networ.},
\textbf{1}(2), 229--232.\label{Luttrell1990b}
\bibitem{AleksanderStonham1979} Aleksander, I., \& Stonham, T. J.
(1979). \textit{Guide to pattern recognition using random access
memories}. \textit{Comput. Digital Techniques}, \textbf{2}, 29--40.
\label{AleksanderStonham1979}
\bibitem{Brodatz1966} Brodatz, P. (1966). \textit{Textures - A Photographic
Album for Artists and Designers}. New York: Dover.\label{Brodatz1966}
\bibitem{Jaynes1968} Jaynes, E. T. (1968). \textit{Prior probabilities}.
\textit{IEEE Trans. Syst. Sci. Cyb.}, \textbf{4}(3), 227--241.\label{Jaynes1968}
\bibitem{Jaynes1982} Jaynes, E. T. (1982). \textit{On the rationale
of maximum entropy methods}. \textit{Proc. IEEE}, \textbf{70}(9),
939--952.\label{Jaynes1982}
\bibitem{Luttrell1989b} Luttrell, S. P. (1989). \textit{Hierarchical
self-organising networks}. In \textit{Proc. IEE Conf. on Artificial
Neural Networks} (pp. 2--6). London: IEE.\label{Luttrell1989b}
\bibitem{Luttrell1989c} Luttrell, S. P. (1989). \textit{Image compression
using a multilayer neural network}. \textit{Patt. Recogn. Lett.},
\textbf{10}(1), 1--7.\label{Luttrell1989c}
\bibitem{LindeBuzoGray1980} Linde, Y., Buzo, A., \& Gray, R. M.
(1980). \textit{An algorithm for vector quantiser design}. \textit{IEEE
Trans. Commun.}, \textbf{28}(1), 84--95.\label{LindeBuzoGray1980}
\bibitem{KumazawaKasaharaNamekawa1984} Kumazawa, H., Kasahara, M.,
\& Namekawa, T. (1984). \textit{A construction of vector quantisers
for noisy channels}. \textit{Electron. Eng. Japan B}, \textbf{67}(4),
39--47.\label{KumazawaKasaharaNamekawa1984}
\bibitem{Farvardin1990} Farvardin, N. (1990). \textit{A study of
vector quantisation for noisy channels}. \textit{IEEE Trans. Inform.
Theory}, \textbf{36}(4), 799--809.\label{Farvardin1990}
\end{thebibliography}
\end{document}